\documentclass{article}

\usepackage{microtype}
\usepackage{graphicx}
\usepackage{subcaption}
\usepackage{booktabs} 

\usepackage{hyperref}

\usepackage[preprint]{icml2026}

\usepackage{times}
\usepackage{soul}
\usepackage{url}
\usepackage[utf8]{inputenc}
\usepackage{graphicx}
\usepackage{amsthm}
\usepackage{booktabs}
\usepackage[switch]{lineno}
\usepackage{tikz-cd}
\usetikzlibrary{arrows}
\usepackage{comment}
\usepackage{amssymb}
\usepackage{tikz}
\usepackage{graphicx} 
\usepackage{tikz-qtree,tikz-qtree-compat}
\usetikzlibrary{positioning,arrows.meta,calc,fit}
\usepackage{subcaption}
\usepackage{enumitem}

\usepackage{mathtools} 
\usepackage{booktabs} 
\usepackage{stackengine}
\usepackage{makecell}
\def\delequal{\mathrel{\ensurestackMath{\stackon[1pt]{=}{\scriptstyle\Delta}}}}

\usepackage[capitalize,noabbrev]{cleveref}

\theoremstyle{plain}
\newtheorem{theorem}{Theorem}[section]

\theoremstyle{definition}
\newtheorem{definition}[theorem]{Definition}

\theoremstyle{remark}

\usepackage[textsize=tiny]{todonotes}

\icmltitlerunning{Submission and Formatting Instructions for ICML 2026}

\begin{document}

\twocolumn[
  \icmltitle{Learning Probabilities of Causation with Mask-Augmented Data}



  \icmlsetsymbol{equal}{*}

  \begin{icmlauthorlist}
    \icmlauthor{Shuai Wang}{fsu}
    \icmlauthor{Yizhou Sun}{ucla}
    \icmlauthor{Judea Pearl}{ucla}
    \icmlauthor{Ang Li}{fsu}

  \end{icmlauthorlist}

  \icmlaffiliation{fsu}{Florida State University, Tallahassee, FL 32306, USA}
  \icmlaffiliation{ucla}{University of California Los Angeles, Los Angeles, CA 90095, USA}

  \icmlcorrespondingauthor{Shuai Wang}{sw23v@cs.fsu.edu}
  \icmlcorrespondingauthor{Yizhou Sun}{yzsun@cs.ucla.edu}
  \icmlcorrespondingauthor{Judea Pearl}{judea@cs.ucla.edu}
  \icmlcorrespondingauthor{Ang Li}{angli@cs.fsu.edu}

  \icmlkeywords{Machine Learning, ICML}

  \vskip 0.3in
]



\printAffiliationsAndNotice{}  

\begin{abstract}
    Probabilities of causation play a central role in modern decision making. Tian and Pearl first introduced formal definitions and derived tight bounds for three binary probabilities of causation, such as the probability of necessity and sufficiency (PNS). However, estimating these probabilities requires both experimental and observational distributions specific to each subpopulation, which are often unreliable or impractical to obtain from limited population-level data. To solve this problem, we propose two machine learning models: Exact-MLP and Mask-MLP, which are trained on a small set of reliable subpopulations and are able to predict PNS bounds for all other subpopulations. We validate our models across four Structural Causal Models (SCMs), each evaluated on population-level data with sample sizes between 100k and 200k. Our models achieve average mean absolute errors (MAEs) of roughly 0.03 on main tasks, reducing MAE by about 80\% relative to the corresponding baselines. These results demonstrate both the feasibility of machine learning models for learning probabilities of causation and the effectiveness of the proposed approach.

\end{abstract}

\section{Introduction}\label{sec:intro}
    Understanding causal relationships and estimating probabilities of causation are crucial in fields such as healthcare, policy evaluation, and economics \cite{pearl2009causality,imbens2015causal,heckman2015causal}. Unlike correlation-based methods, causal inference enables decision-makers to determine whether an action or intervention directly leads to a desired outcome. This is particularly essential in personalized medicine, where accurately assessing treatment effects ensures both efficacy and safety \cite{mueller:pea23-r530}. Moreover, causal reasoning enhances machine learning applications by improving accuracy \cite{li2020training}, interpretability, and fairness \cite{plecko2022causal} in automated decision making. Despite its broad significance, estimating probabilities of causation remains challenging due to data \textbf{limitations}. In this paper, we address this challenge by leveraging machine learning techniques to predict probabilities of causation for subpopulations with \textbf{insufficient} data.
    
    The study of probabilities of causation began around 2000, when researchers introduced three probabilities of causation including PNS within the framework of SCMs \cite{pearl1999probabilities,galles1998axiomatic,halpern2000axiomatizing,pearl2009causality}. In this paper, we focus on PNS, since it is most complex and representative. Specifically, PNS represents the chance that the cause is both required and sufficient for the outcome. And PNS is formally defined in Section~\ref{sec:pre}; its practical value lies in helping us make informed, causality based decisions. For example, while taking Tylenol (a fever reducing drug) is correlated with having a fever, we don’t give Tylenol because of the correlation. Instead, we give them because of a causal belief (PNS): without the drug, fever would likely persist; with the drug, it would likely subside.. Importantly, traditional randomized controlled trials (RCTs), though powerful for estimating average treatment effects, are not always sufficient for answering such counterfactual questions as illustrated in \cite{li2019unit}.
    
    Subsequently, \cite{tian2000probabilities} derived tight bounds for these probabilities using Balke's linear programming \cite{balke1995probabilistic}, incorporating both observational and experimental data\footnote{Observational data refers to data typically collected through surveys, where individuals can access and freely choose whether to receive a treatment. In contrast, experimental data is usually obtained from randomized controlled trials, where individuals are randomly assigned to receive or not receive the treatment.}. Nearly two decades later, \cite{li2019unit} formally proved these bounds and introduced the unit selection model, a decision making framework based on the linear combination of probabilities of causation. More recently, \cite{li2024unit} extended the definitions and bounds to a more general form. Additionally, \cite{pearl:etal21-r505}, as well as \cite{dawid2017}, demonstrated that these bounds could be further refined given specific causal structures.
    
    However, all of the above estimators of probabilities of causation require sufficiently large observational and experimental samples.
    Moreover, following \cite{li2022probabilities}, obtaining reliable estimates for a subpopulation typically needs on the order of $300$ experimental and $300$ observational samples.
    While computing PNS bounds for a small number of well-sampled subpopulations is feasible, producing reliable bounds for \emph{all} subpopulations quickly becomes statistically and computationally intractable.

        Exact subpopulations correspond to fully specified covariate profiles (all covariates are 0 or 1), resulting in $2^{10}{=}1024$ subpopulations.
    Mask-augmented subpopulations allow some covariates to be unspecified (denoted by $X$), yielding $3^{10}{=}59{,}049$ subpopulations.
    We refer to queries over exact subpopulations as \textbf{exact queries}, and queries over mask-augmented subpopulations as \textbf{mask queries}. Note that exact queries are a special case of mask queries.
    
    In this setting, direct plug-in computation of PNS becomes impractical: approximately $17\%$ of mask-augmented subpopulations and $99\%$ of exact subpopulations have no \emph{valid} data after enforcing the per-group sample requirement in \emph{both} datasets, and among the remaining exact subpopulations, the MAE ranges from $0.1499$ to $0.2076$, which is inadequate for practical use.

    Hence, in this research, we will demonstrate the potential of machine learning models to achieve accurate estimations for subpopulations with insufficient data. Our approach first computes PNS bounds for both exact and mask-augmented subpopulations with sufficient data using causal formulas. These bounds are then used to train two Multilayer perceptron (MLP) based predictors \cite{mlp}: the model trained on the mask-augmented query space is called \textbf{Mask-MLP}, while the model trained on exact queries is a standard MLP and is referred to as \textbf{Exact-MLP} for distinction. We evaluate model performance on test data using oracle PNS bounds derived from causal equations under a known data-generating process. To assess prediction quality comprehensively, we evaluate our models across four distinct SCMs and report results averaged over five runs.

    \subsection{Contributions}
    \begin{itemize}[nosep]
        \item \textbf{Learning to predict PNS bounds for subpopulations without enough data:}
        We formalize PNS bounds prediction as a supervised learning problem where only a small subset of subpopulations have sufficient experimental and observational data. Therefore, our models can predict the PNS bounds for all subpopulations.
    
        \item \textbf{Mask-augmented subpopulation modeling that scales supervision from \(2^{d}\) to \(3^{d}\):}
        We introduce a mask-augmented representation of subpopulation queries and train a single \emph{Mask-MLP} on \(3^{d}\) mask-augmented subpopulations. This model not only predicts mask queries, but also achieves better performance on the same exact query prediction task under data scarcity.   
    
    \end{itemize}

\section{Related Work}\label{sec:rel}
    As the demand for explainable AI and causal reasoning grows, recent studies increasingly combine causal inference and machine learning. Broadly, the literature falls into two lines: (i) leveraging causal principles to improve ML models; and (ii) using ML to estimate or extrapolate causal quantities.
    
    \paragraph{Causal principles for ML.}
    By incorporating counterfactual reasoning, \cite{kusner2017counterfactual} proposed a framework for counterfactual fairness.
    A number of explanation methods operationalize the notions of necessity and sufficiency within a predictive model.
    For example, \cite{DBLP:journals/corr/abs-2103-11972} (LEWIS) and \cite{watson2021localexplanationsnecessitysufficiency} (LENS) introduce scores inspired by “necessary/sufficient causes” to highlight influential features.
    However, their operationalizations (e.g., NeSuf) are defined with respect to the \emph{model's} behavior rather than the structural counterfactual semantics of Pearl’s PNS; hence they are not equivalent to PNS under standard SCM definitions.
    These approaches aim primarily at improving explainability of ML predictions rather than estimating causal quantities per se.
    
    \paragraph{ML for causal quantities.}
    \cite{xia2022neuralcausalmodelscounterfactual} (GAN-NCM) learns neural SCMs to approximate counterfactuals from observational and experimental data, but typically assumes access to all relevant features and sufficient data for reliable SCM fitting.
    In contrast, our setting targets \emph{subpopulation-level} probabilities of causation (e.g., PNS bounds) when many subpopulations are sparsely observed or even unobserved.
    Classical identification and bounding results (e.g., Tian–Pearl bounds and related LP formulations) provide analytic or optimization-based constraints.
    Our contribution is complementary: given reliable bounds on a small set of subpopulations, we learn to extrapolate \emph{feasible and calibrated} PNS bounds to long-tail subpopulations under data scarcity, and evaluate them by feasibility, coverage, and policy-oriented ranking metrics.

\section{Preliminaries}\label{sec:pre}
    In this section, we review the fundamental concepts of causal inference necessary for understanding the rest of the paper. We begin by discussing the definition of PNS as introduced by \cite{pearl1999probabilities}, followed by the tight bounds of PNS \cite{tian2000probabilities}. Experienced readers may skip this section.
    
    Similar to the works mentioned above, we adopt the causal language of Structural Causal Models (SCMs) \cite{galles1998axiomatic,halpern2000axiomatizing}. In this framework, the counterfactual statement ``Variable \( Y \) would have the value \( y \) had \( X \) been \( x \)'' is denoted as \( Y_x = y \), abbreviated as \( y_x \). We consider two types of data: experimental data, expressed as causal effects \( P(y_x) \), and observational data, represented by the joint probability function \( P(x, y) \). Unless otherwise specified, we assume \( X \) and \( Y \) are binary variables in a causal model \( M \), with \( x \) and \( y \) denoting the propositions \( X = \text{true} \) and \( Y = \text{true} \), respectively, and \( x' \) and \( y' \) representing their complements. For simplicity, we focus on binary treatments and effects; extensions to multi-valued cases are discussed by \cite{pearl2009causality} (p. 286, footnote 5) and \cite{li2024probabilities}.
    
    First, the definition of PNS, is defined using SCM as follow \cite{pearl1999probabilities}:
    
    \begin{definition}[Probability of necessity and sufficiency (PNS)] Let $X$ and $Y$ be two binary variables in a causal model $M$, let $x$ and $y$ stand for the propositions $X=true$ and $Y=true$, respectively, and $x'$ and $y'$ for their complements. The probability of necessity and sufficiency is defined as the expression
    \begin{eqnarray}
    \text{PNS} &\delequal& P(Y_{X=true}=true,Y_{X=false}=false)\nonumber\\
    &\delequal& P(y_x,y'_{x'}) \nonumber.
    \end{eqnarray}
    \end{definition}
    Then, the oracle bounds of PNS is given by \cite{tian2000probabilities}.
    \begin{eqnarray}
    \max \left \{
    \begin{array}{cc}
    0, \\
    P(y_x) - P(y_{x'}), \\
    P(y) - P(y_{x'}), \\
    P(y_x) - P(y)
    \end{array}
    \right \} \le
    \text{PNS,}\label{pnslb}\\
    \min \left \{
    \begin{array}{cc}
     P(y_x), \\
     P(y'_{x'}), \\
    P(x,y) + P(x',y'), \\
    P(y_x) - P(y_{x'}) +\\
    P(x, y') + P(x', y)
    \end{array} 
    \right \}\ge
    \text{PNS.}
    \label{pnsub}
    \end{eqnarray}
    
    The advantage of PNS, as illustrated by \cite{li2019unit}, lies in capturing the portion of effects attributable specifically to the treatment, rather than the total causal effect. The primary objective of this paper is then to predict Equations \ref{pnslb} and \ref{pnsub} (i.e., the lower and upper bounds of the PNS) for any subpopulations using those with sufficient data (i.e., sufficient data to estimate the distributions $P(X,Y)$ and $P(Y_X)$.) Due to space constraints, the focus will be on the bounds of PNS (i.e., Equation \ref{pnslb} and \ref{pnsub}). Unless otherwise specified, the discussion will be limited to binary treatment and effect, meaning both $X$ and $Y$ are binary.

\section{Main Framework}\label{sec:method}

    \subsection{The Machine Learning Pipeline}
    Our framework consists of two stages, summarized in Figure~\ref{fig:pipeline}.  Firstly,  from an SCM we compute oracle PNS bounds via Eqs.~\eqref{pnslb}--\eqref{pnsub} as test label, and derive sample-based bounds from finite experimental/observational data (after threshold filtering) as training labels. We then train predictors under either exact supervision on \(q\in\{0,1\}^d\) or mask-augmented supervision on \(q\in\{0,1,X\}^d\), and evaluate by MAE against the oracle bounds on exact and masked queries.
    \begin{figure*}[t]
    \centering
    \resizebox{\textwidth}{!}{%
    \begin{tikzpicture}[
      font=\small,
      box/.style={draw, rounded corners, align=center, inner sep=7pt, minimum height=9mm},
      group/.style={draw, rounded corners, inner sep=10pt},
      arrow/.style={-Latex, thick}
    ]
    
    \node[box, minimum width=18mm] (scm) {SCM};
    
    \node[coordinate, right=16mm of scm] (laneTopStart) {};
    \node[coordinate, below=18mm of laneTopStart] (laneBotStart) {};
    
    \node[box, right=16mm of scm, yshift=9mm, minimum width=62mm] (oracle)
    {Oracle Test Set\\
    Queries $q\in\{0,1,X\}^{d}$\\
    $y^{\star}(q)$};
    
    \node[box, below=10mm of oracle, minimum width=62mm] (finite)
    {Finite Exp/Obs Samples\\
    $\Rightarrow$ Sample-based Bounds\\
    $[\widehat{LB}(q),\,\widehat{UB}(q)]$\\
    (Filtered by threshold)};
    
    \node[group, fit=(oracle)(finite)] (dg) {};
    \node[font=\small\bfseries, fill=white, inner sep=1.5pt]
      at ($(dg.north)+(0,2mm)$) {Data Generation};
    
    \draw[arrow] (scm.east) -- node[above, align=center] {compute\\oracle} (oracle.west);
    \draw[arrow] (scm.east) -- node[below, align=center] {sample\\exp+obs} (finite.west);
    
    \node[box, right=22mm of oracle, minimum width=58mm] (exacttrain)
    {Exact Supervision\\
    $q\in\{0,1\}^{d}$ \quad size $2^{d}$};
    
    \node[box, right=16mm of exacttrain, minimum width=58mm] (exactmlp)
    {Exact-MLP\\
    MLP (Mish)\\
    train on $2^{d}$, predict exact};
    
    \node[box, right=16mm of exactmlp, minimum width=44mm] (exactout)
    {Predicted\\$[\widehat{LB},\widehat{UB}]$\\on exact};
    
    \draw[arrow] (finite.east) -- ++(8mm,0) |- (exacttrain.west);
    \draw[arrow] (exacttrain.east) -- (exactmlp.west);
    \draw[arrow] (exactmlp.east) -- (exactout.west);
    
    \draw[arrow] (oracle.east) -- ++(8mm,0) |- (exactmlp.north);
    
    \node[box, right=22mm of finite, minimum width=58mm] (masktrain)
    {Mask-aug Supervision\\
    $q\in\{0,1,X\}^{d}$ \quad size $3^{d}$};
    
    \node[box, right=16mm of masktrain, minimum width=58mm] (maskmlp)
    {Mask-MLP\\
    MLP (Mish)\\
    train on $3^{d}$, predict masked \& exact};
    
    \node[box, right=16mm of maskmlp, minimum width=44mm] (maskout)
    {Predicted\\$[\widehat{LB},\widehat{UB}]$\\on masked\\(and exact subset)};
    
    \draw[arrow] (finite.east) -- ++(8mm,0) |- (masktrain.west);
    \draw[arrow] (masktrain.east) -- (maskmlp.west);
    \draw[arrow] (maskmlp.east) -- (maskout.west);
    
    \draw[arrow] (oracle.east) -- ++(8mm,0) |- (maskmlp.north);
    
    \node[group, fit=(exacttrain)(masktrain)] (tp) {};
    \node[font=\small\bfseries, fill=white, inner sep=1.5pt]
      at ($(tp.north)+(0,2mm)$) {Training Protocols};
    
    \node[group, fit=(exactmlp)(maskmlp)] (ml) {};
    \node[font=\small\bfseries, fill=white, inner sep=1.5pt]
      at ($(ml.north)+(0,2mm)$) {Machine Learning};
    
    \end{tikzpicture}%
    }
    \caption{Pipeline for predicting PNS bounds under exact and mask queries. Oracle test labels are computed from the SCM, while training labels are sample-based bounds estimated from finite experimental and observational data.}
    \label{fig:pipeline}
    \end{figure*}

    \subsection{Causal Dataset Generation}
    \begin{figure}[t]
        \centering
        \begin{subfigure}[b]{0.2\textwidth}
            \centering
            \begin{tikzpicture}[->,>=stealth',node distance=2cm,
              thick,main node/.style={circle,fill,inner sep=1.5pt}]
              \node[main node] (1) [label=above:{$Z$}]{};
              \node[main node] (2) [below left =1cm of 1,label=left:$X$]{};
              \node[main node] (3) [below right =1cm of 1,label=right:$Y$] {};
              \path[every node/.style={font=\sffamily\small}]
                (1) edge node {} (2)
                (1) edge node {} (3)
                (2) edge node {} (3);
            \end{tikzpicture}
            \caption{Confounder $Z$ with direct effect}
            \label{fig:confounder}
        \end{subfigure}
        \hfill
        \begin{subfigure}[b]{0.2\textwidth}
            \centering
            \begin{tikzpicture}[->,>=stealth',node distance=2cm,
              thick,main node/.style={circle,fill,inner sep=1.5pt}]
              \node[main node] (1) [label=above:{$Z$}]{};
              \node[main node] (2) [below left =1cm of 1,label=left:$X$]{};
              \node[main node] (3) [below right =1cm of 1,label=right:$Y$] {};
              \path[every node/.style={font=\sffamily\small}]
                (1) edge node {} (3)
                (2) edge node {} (3);
            \end{tikzpicture}
            \caption{Covariate $Z$}
            \label{fig:outcome covariate}
        \end{subfigure}
        \hfill
        \begin{subfigure}[b]{0.2\textwidth}
            \centering
            \begin{tikzpicture}[->,>=stealth',node distance=2cm,
              thick,main node/.style={circle,fill,inner sep=1.5pt}]
              \node[main node] (1) [label=above:{$Z$}]{};
              \node[main node] (2) [below left =1cm of 1,label=left:$X$]{};
              \node[main node] (3) [below right =1cm of 1,label=right:$Y$] {};
              \path[every node/.style={font=\sffamily\small}]
                (2) edge node {} (3);
            \end{tikzpicture}
            \caption{Unrelated $Z$ with Direct effect}
            \label{fig:direct effect}
        \end{subfigure}
        \hfill
        \begin{subfigure}[b]{0.2\textwidth}
            \centering
            \begin{tikzpicture}[->,>=stealth',node distance=1.4cm,
              thick,main node/.style={circle,fill,inner sep=1.5pt}]
              \node[main node] (1) [label=above:{$Z$}]{};
              \node[main node] (2) [below left =0.4cm and 0.7cm of 1,label=left:$X$]{};
              \node[main node] (3) [below right =0.4cm and 0.7cm of 1,label=right:$Y$]{};
              \node[main node] (4) [below =0.8cm of 1,label=below:$M$]{};
    
              \path[every node/.style={font=\sffamily\small}]
                (1) edge (2)
                (1) edge (3)
                (2) edge (3)
                (2) edge (4)
                (4) edge (3);
            \end{tikzpicture}
            \caption{Mediator $M$ and direct effect}
            \label{fig:mediator_with_direct}
        \end{subfigure}
    
        \caption{Different SCMs in this study.}
        \label{fig:different_scm}
    \end{figure}

    \textbf{SCMs and subpopulation queries.}
    We generate synthetic data from four SCMs shown in Figure~\ref{fig:different_scm} (full structural equations and additional details are provided in the appendix).
    Our goal is to predict PNS bounds for both exact and mask queries defined over $d$ observed binary covariates.
    Exact queries $q\in\{0,1\}^{d}$ fully specify a covariate profile, whereas mask queries take the form $q\in\{0,1,X\}^{d}$, where $X$ denotes an unspecified covariate value.
    For example, $q=(1,X,0)$ represents the union of the two exact profiles obtained by setting the second entry to $0$ or $1$.
    Exact queries are a special case of mask-augmented queries.
    
    \vspace{0.5em}
    \textbf{Oracle test set.}
    To evaluate model with accurate bounds, we compute \emph{oracle} causal quantities from SCM by causal equation.
    For any query $q$, we compute oracle experimental probability $P(y_x\mid q)$ and $P(y_{x'}\mid q)$ as well as the oracle observational joint distribution $P(x,y\mid q)$ by marginalizing over all hidden variables in the SCM (details in the appendix).
    These oracle distributions are then plugged into the Tian--Pearl bounds in Eqs.~\eqref{pnslb}--\eqref{pnsub} to obtain the oracle lower and upper bounds of PNS for query $q$, which serve as the \textbf{test labels}.
    And a mask query $q\in\{0,1,X\}^{d}$ corresponds to a collection of fully specified covariate vectors $z\in\{0,1\}^{d}$ obtained by filling in the masked entries (those with $X$).
    Let $\mathcal{Z}(q)$ denote the set of such exact covariate vectors consistent with $q$, and define the normalizer

    \begin{equation}
    \pi(q) := \sum_{z\in\mathcal{Z}(q)} P(z).
    \label{eq:pi_q}
    \end{equation}
    Oracle quantities conditioned on $q$ are computed by mixture aggregation over $\mathcal{Z}(q)$:
    \begin{equation}
    \begin{aligned}
    P(y_x \mid q)
    &= \frac{1}{\pi(q)}\sum_{z\in\mathcal{Z}(q)} P(y_x\mid z)\,P(z),\\
    P(x,y \mid q)
    &= \frac{1}{\pi(q)}\sum_{z\in\mathcal{Z}(q)} P(x,y\mid z)\,P(z).
    \end{aligned}
    \label{eq:oracle_mix_mask_main}
    \end{equation}
    The same aggregation applies to $P(y_{x'}\mid q)$.
    Finally, we obtain oracle PNS bounds for $q$ by plugging the aggregated oracle quantities into Eqs.~\eqref{pnslb}--\eqref{pnsub}.

    \vspace{0.5em}
    \textbf{Generate samples for training labels.}
    To emulate realistic data scarcity, we need to generate a finite number of observational and experimental samples from the SCM.
    For the observational setting, $(X,Y)$ are generated according to the causal equations.
    And for the experimental setting, $X$ is randomized and the remaining  variables are generated by the causal equations.
    Only observed covariates and outcomes are retained; all hidden variables are unobserved. Note that all uncertainty are caused by those unobserved variables, which means if all variables are observed, the PNS would be an accurate value rather than bounds.
    
    \vspace{0.5em}
    \textbf{Calculate bounds for training samples}
    Training labels are computed from finite samples by estimating. For each query $q$, we obtain the experimental distribution $P(y_x\mid q)$ and $P(y_{x'}\mid q)$ from the experimental samples and the observational distribution $P(x,y\mid q)$ from the observational samples. Then we plugging these empirical estimates into Eqs.~\eqref{pnslb}--\eqref{pnsub}.
    To ensure feasibility, we add a query $q$ into the training set only if it has sufficient support in \emph{both} experimental and observational samples (e.g., at least $300$ samples each \cite{li2022probabilities}) with constraints (e.g., $\mathrm{LB}\le \mathrm{UB}$).
    This yields two supervised datasets:
    (i) an \textbf{exact} training set over $q\in\{0,1\}^{d}$ (used for Exact-MLP), and
    (ii) a \textbf{mask-augmented} training set over $q\in\{0,1,X\}^{d}$ (used for Mask-MLP).

    \subsection{Machine Learning}
    We study whether a multilayer perceptron (MLP) can predict PNS bounds to subpopulation queries with insufficient data. Given the considerable number of zero values in the data, we proposed using $\text{Mish}(s) = s \cdot \tanh(\ln(1 + e^s))$ \cite{mish}, as a smooth and stable activation function rather than ReLU, to train two independent regressors for the lower and upper bounds.
    
    \paragraph{Input representation.}
    Each query $q\in\{0,1,X\}^{d}$ is encoded by a $3d$-dimensional one-hot vector.
    For each coordinate $j\in[d]$ and each symbol $c\in\{0,1,X\}$, define
    \begin{equation}
    \phi_{j,c}(q)=\mathbb{I}[q_j=c].
    \label{eq:phi_def}
    \end{equation}
    We then form the feature vector $\phi(q)\in\{0,1\}^{3d}$ by concatenating these indicators over all $(j,c)$ pairs in a fixed order:
    \begin{equation}
    \phi(q)=\mathrm{concat}\big(\{\phi_{j,c}(q)\}_{j\in[d],\,c\in\{0,1,X\}}\big).
    \label{eq:phi_concat}
    \end{equation}
    
    \paragraph{Predictors.}
    We train two independent MLP regressors, $f^{\mathrm{LB}}$ and $f^{\mathrm{UB}}$, to predict the lower and upper bounds, respectively.
    Both networks share the same architecture and use Mish activations; the output layer uses a sigmoid so that predictions lie in $[0,1]$.
    The training data are the sample-based bounds computed from finite experimental and observational data (Section~\ref{sec:method}).
    
    \paragraph{Training protocols.}
    \textbf{Exact-MLP (exact supervision).}
    We train $f^{\mathrm{LB}}$ and $f^{\mathrm{UB}}$ on exact queries only, i.e., $q\in\{0,1\}^{d}$ (up to $2^{d}$ queries after thresholding), and evaluate on oracle bounds for exact queries.
    
    \textbf{Mask-MLP (mask-augmented supervision).}
    We train the same architecture on the enlarged query space $q\in\{0,1,X\}^{d}$ (up to $3^{d}$ queries after thresholding).
    The resulting model is evaluated on oracle bounds for both mask queries and the exact-query subset.
    
    \paragraph{Evaluation.}
    We compare the predicted bounds to oracle PNS bounds computed from the SCM and report the average MAEs for both the lower and upper bounds. Since the threshold setting affects the model MAEs, to evaluate different models fairly, we test the thresholds from 100 to 2000, and select the best MAE and its corresponding threshold to represent the performance. The reported MAEs are averaged across five runs.

    \subsection{Algorithm}
    Through the whole process, we can summarize the process into two algorithms as \ref{alg:Exact-MLP} and \ref{alg:Mask-MLP}. In a word, for each threshold $\tau$, the algorithm will calculate the oracle PNS bounds and sample-based bounds by formula \ref{pnslb} and \ref{pnsub}. Then the PNS bounds would be added to training data $\mathcal{D}_\tau$ iff this subgroup contains enough legal samples (larger than $\tau$). Next, we train the MLP $f^{LB}_\tau$ and $f^{UB}_\tau$, and obtain the MAEs of the MLP compared with the oracle PNS bounds. Finally, we obtain the MAEs through different thresholds $\tau$.

    \begin{algorithm}[t]
        \caption{Exact-MLP}
        \label{alg:Exact-MLP}
        \small
        \begin{algorithmic}[1]
        \REQUIRE SCM $M$ with $d$ observed binary covariates; budgets $(N_{\text{exp}},N_{\text{obs}})$; threshold grid $\mathcal{T}$.
        \ENSURE Predictors and MAE on oracle PNS bounds over $\mathcal{Q}_{\text{exact}}=\{0,1\}^{d}$.
        
        \STATE Sample experimental data $\mathcal{D}_{\text{exp}}$ (size $N_{\text{exp}}$) and observational data $\mathcal{D}_{\text{obs}}$ (size $N_{\text{obs}}$) from $M$.
        \STATE Compute oracle bounds $\{(LB^\star(q),UB^\star(q))\}_{q\in\mathcal{Q}_{\text{exact}}}$ via Eqs.~\eqref{pnslb}--\eqref{pnsub}.
        
        \FOR{$\tau \in \mathcal{T}$}
          \STATE Initialize training set $\mathcal{D}_\tau \leftarrow \emptyset$.
        
          \FOR{$q \in \mathcal{Q}_{\text{exact}}$}
            \STATE Estimate experimental distribution on $q$ from $\mathcal{D}_{\text{exp}}$:
            \STATE \hspace*{1em}$\widehat{P}(y_x\!\mid q)=\widehat{P}(Y{=}1 \mid do(X{=}1),q)$
            \STATE \hspace*{1em}$\widehat{P}(y_{x'}\!\mid q)=\widehat{P}(Y{=}1 \mid do(X{=}0),q)$
        
            \STATE Estimate the observational distribution table on $q$ from $\mathcal{D}_{\text{obs}}$:
            \STATE \hspace*{1em}$\widehat{P}(X,Y\mid q)$ (the full $2\times2$ table over $X,Y\in\{0,1\}$).
        
            \STATE Compute sample-based bounds $[\widehat{LB}_\tau(q),\widehat{UB}_\tau(q)]$ via Eqs.~\eqref{pnslb}--\eqref{pnsub}.
        
            \STATE Compute supports $\text{support}_{\text{exp}}(q)$ in $\mathcal{D}_{\text{exp}}$ and $\text{support}_{\text{obs}}(q)$ in $\mathcal{D}_{\text{obs}}$.
        
            \IF{$\text{support}_{\text{exp}}(q)\ge \tau$ \AND $\text{support}_{\text{obs}}(q)\ge \tau$ \AND $0\le \widehat{LB}_\tau(q)\le \widehat{UB}_\tau(q)\le 1$}
              \STATE Add $(\phi(q),\widehat{LB}_\tau(q),\widehat{UB}_\tau(q))$ to $\mathcal{D}_\tau$.
            \ENDIF
          \ENDFOR
        
          \STATE Train two regressors $f^{LB}_\tau$ and $f^{UB}_\tau$ (MLP+Mish) on $\mathcal{D}_\tau$.
          \STATE Predict for all $q\in\mathcal{Q}_{\text{exact}}$:
          \STATE \hspace*{1em}$\widehat{LB}(q)=f^{LB}_\tau(\phi(q))$, \ $\widehat{UB}(q)=f^{UB}_\tau(\phi(q))$.
        
          \STATE Record
          $\mathrm{MAE}_{LB}(\tau)=\mathbb{E}_{q\in\mathcal{Q}_{\text{exact}}}\!\left|\widehat{LB}(q)-LB^\star(q)\right|$,
          $\mathrm{MAE}_{UB}(\tau)=\mathbb{E}_{q\in\mathcal{Q}_{\text{exact}}}\!\left|\widehat{UB}(q)-UB^\star(q)\right|$.
        \ENDFOR
        
        \STATE Report the best $\tau$ (or the full sweep over $\mathcal{T}$) and the corresponding predictors.
        \end{algorithmic}
    \end{algorithm}

    \begin{algorithm}[t]
        \caption{Mask-MLP}
        \label{alg:Mask-MLP}
        \small
        \begin{algorithmic}[1]
        \REQUIRE SCM $M$ with $d$ observed binary covariates; budgets $(N_{\text{exp}},N_{\text{obs}})$; threshold grid $\mathcal{T}$.
        \ENSURE Predictors and MAE on oracle PNS bounds over $\mathcal{Q}_{\text{mask}}=\{0,1,X\}^{d}$ and $\mathcal{Q}_{\text{exact}}\subset\mathcal{Q}_{\text{mask}}$.
        
        \STATE Sample experimental data $\mathcal{D}_{\text{exp}}$ (size $N_{\text{exp}}$) and observational data $\mathcal{D}_{\text{obs}}$ (size $N_{\text{obs}}$) from $M$.
        \STATE Compute oracle bounds $\{(LB^\star(q),UB^\star(q))\}_{q\in\mathcal{Q}_{\text{mask}}}$ via Eqs.~\eqref{pnslb}--\eqref{pnsub}.
        \STATE For masked queries, aggregate oracle quantities as in Eq.~\eqref{eq:oracle_mix_mask_main}.
        
        \FOR{$\tau \in \mathcal{T}$}
          \STATE Initialize training set $\mathcal{D}_\tau \leftarrow \emptyset$.
        
          \FOR{$q \in \mathcal{Q}_{\text{mask}}$}
            \STATE Estimate experimental distribution on $q$ from $\mathcal{D}_{\text{exp}}$:
            \STATE \hspace*{1em}$\widehat{P}(y_x\!\mid q)=\widehat{P}(Y{=}1 \mid do(X{=}1),q)$
            \STATE \hspace*{1em}$\widehat{P}(y_{x'}\!\mid q)=\widehat{P}(Y{=}1 \mid do(X{=}0),q)$
        
            \STATE Estimate the observational distribution table on $q$ from $\mathcal{D}_{\text{obs}}$:
            \STATE \hspace*{1em}$\widehat{P}(X,Y\mid q)$ (the full $2\times2$ table).
            \STATE For masked $q$, obtain these estimates by aggregating exact-profile estimates as in Eq.~\eqref{eq:oracle_mix_mask_main}.
        
            \STATE Compute sample-based bounds $[\widehat{LB}_\tau(q),\widehat{UB}_\tau(q)]$ via Eqs.~\eqref{pnslb}--\eqref{pnsub}.
            \STATE Compute supports $\text{support}_{\text{exp}}(q)$ and $\text{support}_{\text{obs}}(q)$.
        
            \IF{$\text{support}_{\text{exp}}(q)\ge \tau$ \AND $\text{support}_{\text{obs}}(q)\ge \tau$ \AND $0\le \widehat{LB}_\tau(q)\le \widehat{UB}_\tau(q)\le 1$}
              \STATE Add $(\phi(q),\widehat{LB}_\tau(q),\widehat{UB}_\tau(q))$ to $\mathcal{D}_\tau$.
            \ENDIF
          \ENDFOR
        
          \STATE Train two regressors $g^{LB}_\tau$ and $g^{UB}_\tau$ (MLP+Mish) on $\mathcal{D}_\tau$.
          \STATE Predict for all $q\in\mathcal{Q}_{\text{mask}}$:
          \STATE \hspace*{1em}$\widehat{LB}(q)=g^{LB}_\tau(\phi(q))$, \ $\widehat{UB}(q)=g^{UB}_\tau(\phi(q))$.
        
          \STATE Record MAEs on both spaces:
          \STATE \hspace*{1em}$\mathrm{MAE}^{\text{mask}}_{LB}(\tau)=\mathbb{E}_{q\in\mathcal{Q}_{\text{mask}}}\!\left|\widehat{LB}(q)-LB^\star(q)\right|$
          \STATE \hspace*{1em}$\mathrm{MAE}^{\text{mask}}_{UB}(\tau)=\mathbb{E}_{q\in\mathcal{Q}_{\text{mask}}}\!\left|\widehat{UB}(q)-UB^\star(q)\right|$
          \STATE \hspace*{1em}$\mathrm{MAE}^{\text{exact}}_{LB}(\tau)=\mathbb{E}_{q\in\mathcal{Q}_{\text{exact}}}\!\left|\widehat{LB}(q)-LB^\star(q)\right|$
          \STATE \hspace*{1em}$\mathrm{MAE}^{\text{exact}}_{UB}(\tau)=\mathbb{E}_{q\in\mathcal{Q}_{\text{exact}}}\!\left|\widehat{UB}(q)-UB^\star(q)\right|$
        \ENDFOR
        
        \STATE Report the best $\tau$ (or the full sweep over $\mathcal{T}$) and the corresponding predictors.
        \end{algorithmic}
    \end{algorithm}

\section{Experimental Results}\label{sec:experiment}

    \subsection{Experimental Setup}
    We evaluate on four SCMs (Confounder, Covariate, Direct, and Mediator) as in Figure~\ref{fig:different_scm}.
    The number of observed covariates $d$ is set to $10$.
    Thereby, we have such two query spaces: (i) exact queries $q\in\{0,1\}^{d}$ (size $2^{10}=1024$), and
    (ii) mask queries $q\in\{0,1,X\}^{d}$ (size $3^{10}=59{,}049$), where $X$ denotes an unspecified covariate.
    Oracle PNS bounds are computed by using the Tian--Pearl formulas in Eqs.~\eqref{pnslb}--\eqref{pnsub}.
    For mask queries, oracle quantities are aggregated as in Eq.~\eqref{eq:oracle_mix_mask_main} before applying Eqs.~\eqref{pnslb}--\eqref{pnsub}.
    To evaluate the relation between data budget and model performance, we generate finite experimental and observational samples under budgets from 50k to 500k and construct sample-based training labels after threshold filtering (Section~\ref{sec:method}).
    We run five independent trials (random seeds 1--5) and report MAE for both the lower and upper bounds.
    Our main predictors are Exact-MLP (trained on exact subpopulations) and Mask-MLP (trained on mask-augmented subpopulations), both using the same MLP+Mish backbone (Section~\ref{sec:method}). While baseline computes sample-based bounds directly by Eqs.~\eqref{pnslb}--\eqref{pnsub} for every possible subpopulation, and then remove bounds that are not feasible.
    
    \begin{table*}[!t]
        \centering
        \caption{MAE on oracle PNS bounds for exact queries $q\in\{0,1\}^{d}$ and mask queries $q\in\{0,1,X\}^{d}$ (Mask-MLP only). Boldface marks the lowest MAE in each panel (LB/UB).}

        \label{tab:combined_results}
    
        \footnotesize
        \setlength{\tabcolsep}{4.5pt}
        \renewcommand{\arraystretch}{1.12}
    
        \begin{tabular}{l l  c c  c c  c c  c c  c c}
        \toprule
         &  &
         \multicolumn{6}{c}{\makecell[c]{\textbf{Exact queries}\\$q\in\{0,1\}^d$}}
         & \multicolumn{4}{c}{\makecell[c]{\textbf{Mask queries}\\$q\in\{0,1,X\}^d$}} \\
        \cmidrule(lr){3-8}\cmidrule(lr){9-12}
    
        \textbf{SCM} & \textbf{Budget}
        & \multicolumn{2}{c}{\textbf{Baseline}}
        & \multicolumn{2}{c}{\makecell[c]{\textbf{Exact-MLP}\\\textbf{(Mish)}}}
        & \multicolumn{2}{c}{\makecell[c]{\textbf{Mask-MLP}\\\textbf{(Mish)}}}
        & \multicolumn{2}{c}{\textbf{Baseline}}
        & \multicolumn{2}{c}{\makecell[c]{\textbf{Mask-MLP}\\\textbf{(Mish)}}} \\
        \cmidrule(lr){3-4}\cmidrule(lr){5-6}\cmidrule(lr){7-8}\cmidrule(lr){9-10}\cmidrule(lr){11-12}
    
        & & \textbf{LB} & \textbf{UB}
          & \textbf{LB} & \textbf{UB}
          & \textbf{LB} & \textbf{UB}
          & \textbf{LB} & \textbf{UB}
          & \textbf{LB} & \textbf{UB} \\
        \midrule
    
        Direct & 50k  & 0.1694 & 0.2089 & 0.0207 & 0.0255 & \textbf{0.0188} & \textbf{0.0247} & 0.0959 & 0.1168 & \textbf{0.0139} & \textbf{0.0179} \\
        Direct & 100k & 0.1499 & 0.1786 & \textbf{0.0180} & \textbf{0.0181} & 0.0196 & 0.0234 & 0.0859 & 0.0987 & \textbf{0.0145} & \textbf{0.0159} \\
        Direct & 200k & 0.1546 & 0.1885 & \textbf{0.0199} & 0.0227 & 0.0218 & \textbf{0.0222} & 0.0754 & 0.0809 & \textbf{0.0145} & \textbf{0.0153} \\
        Direct & 500k & 0.1268 & 0.1540 & \textbf{0.0112} & \textbf{0.0096} & 0.0204 & 0.0197 & 0.0606 & 0.0710 & \textbf{0.0129} & \textbf{0.0129} \\
        \midrule
    
        Covariate & 50k  & 0.1958 & 0.2388 & 0.0623 & 0.0812 & \textbf{0.0375} & \textbf{0.0507} & 0.1095 & 0.1405 & \textbf{0.0240} & \textbf{0.0294} \\
        Covariate & 100k & 0.2076 & 0.1867 & 0.0606 & 0.0772 & \textbf{0.0308} & \textbf{0.0360} & 0.1157 & 0.1036 & \textbf{0.0202} & \textbf{0.0221} \\
        Covariate & 200k & 0.1754 & 0.1825 & 0.0434 & 0.0630 & \textbf{0.0261} & \textbf{0.0377} & 0.0922 & 0.0937 & \textbf{0.0160} & \textbf{0.0223} \\
        Covariate & 500k & 0.1541 & 0.1567 & \textbf{0.0232} & 0.0379 & 0.0287 & \textbf{0.0299} & 0.0750 & 0.0733 & \textbf{0.0172} & \textbf{0.0172} \\
        \midrule
    
        Confounder & 50k  & 0.2166 & 0.2423 & 0.0572 & 0.0874 & \textbf{0.0463} & \textbf{0.0667} & 0.1337 & 0.1392 & \textbf{0.0262} & \textbf{0.0388} \\
        Confounder & 100k & 0.2072 & 0.2101 & 0.0575 & 0.0819 & \textbf{0.0380} & \textbf{0.0500} & 0.1165 & 0.1225 & \textbf{0.0238} & \textbf{0.0288} \\
        Confounder & 200k & 0.1831 & 0.1945 & 0.0275 & 0.0689 & \textbf{0.0251} & \textbf{0.0428} & 0.0992 & 0.1039 & \textbf{0.0164} & \textbf{0.0257} \\
        Confounder & 500k & 0.1523 & 0.1786 & \textbf{0.0244} & \textbf{0.0382} & 0.0307 & 0.0399 & 0.0752 & 0.0800 & \textbf{0.0182} & \textbf{0.0219} \\
        \midrule
    
        Mediator & 50k  & 0.2089 & 0.2256 & 0.0849 & 0.1323 & \textbf{0.0651} & \textbf{0.0725} & 0.1230 & 0.1302 & \textbf{0.0410} & \textbf{0.0483} \\
        Mediator & 100k & 0.1917 & 0.2027 & 0.0731 & 0.1286 & \textbf{0.0441} & \textbf{0.0506} & 0.1175 & 0.1115 & \textbf{0.0282} & \textbf{0.0306} \\
        Mediator & 200k & 0.1838 & 0.1873 & 0.0500 & 0.0607 & \textbf{0.0427} & \textbf{0.0365} & 0.0992 & 0.0878 & \textbf{0.0249} & \textbf{0.0223} \\
        Mediator & 500k & 0.1564 & 0.1472 & \textbf{0.0273} & 0.0423 & 0.0301 & \textbf{0.0221} & 0.0755 & 0.0699 & \textbf{0.0184} & \textbf{0.0132} \\
        \bottomrule
        \end{tabular}
    \end{table*}
    
    \subsection{Overall Accuracy on Exact and Mask Queries}
    Table~\ref{tab:combined_results} summarizes MAE under both query spaces:
    the left panel reports performance on exact queries $q\in\{0,1\}^{d}$, and the right panel reports performance on the mask-augmented space $q\in\{0,1,X\}^{d}$.
    Across all SCMs and budgets, both MLP models substantially outperform the plug-in baseline.
    Overall, Mask-MLP is typically strongest on the exact subset as well, which indicates that training on the enlarged $3^{d}$ query space provides additional information and regularization. However, if we compare Exact-MLP with Mask-MLP on exact queries, there are two exceptions. The first occurs in the simplest Direct structure: Because the PNS bounds are fixed in the Direct structure, fewer training samples suffice, and Exact-MLP performs better for budgets above 100k. Second, their performances are close when the data budget is 500k, because this size can provide nearly enough training data for Exact-MLP.
    On the mask-augmented space, only Mask-MLP applies directly and it consistently improves over the baseline across SCMs and budgets, demonstrating that mask augmentation substantially expands the supported query space while retaining strong accuracy. In summary, both MLP models substantially outperform the baseline. Moreover, Mask-MLP is more robust under data scarcity and can predict PNS bounds for all mask-augmented subpopulations.

    \subsection{Threshold Sensitivity and Stability}
\begin{figure*}[!t]
    \centering
    \setlength{\tabcolsep}{2pt}     
    \renewcommand{\arraystretch}{0} 

    \begin{tabular}{@{}cc@{}}
        \begin{minipage}[b]{0.5\textwidth}
            \centering
            \includegraphics[width=0.99\linewidth]{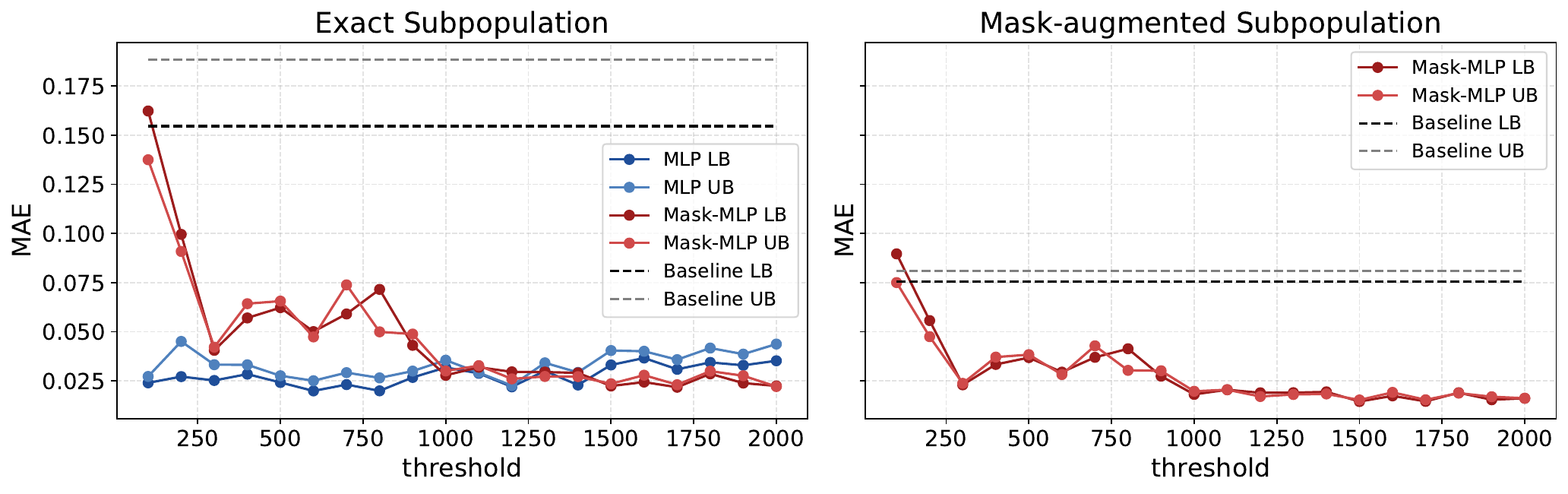}
            \subcaption{Direct (200k)}
            \label{fig:MAE_direct_200k}
        \end{minipage}
        &
        \begin{minipage}[b]{0.5\textwidth}
            \centering
            \includegraphics[width=0.99\linewidth]{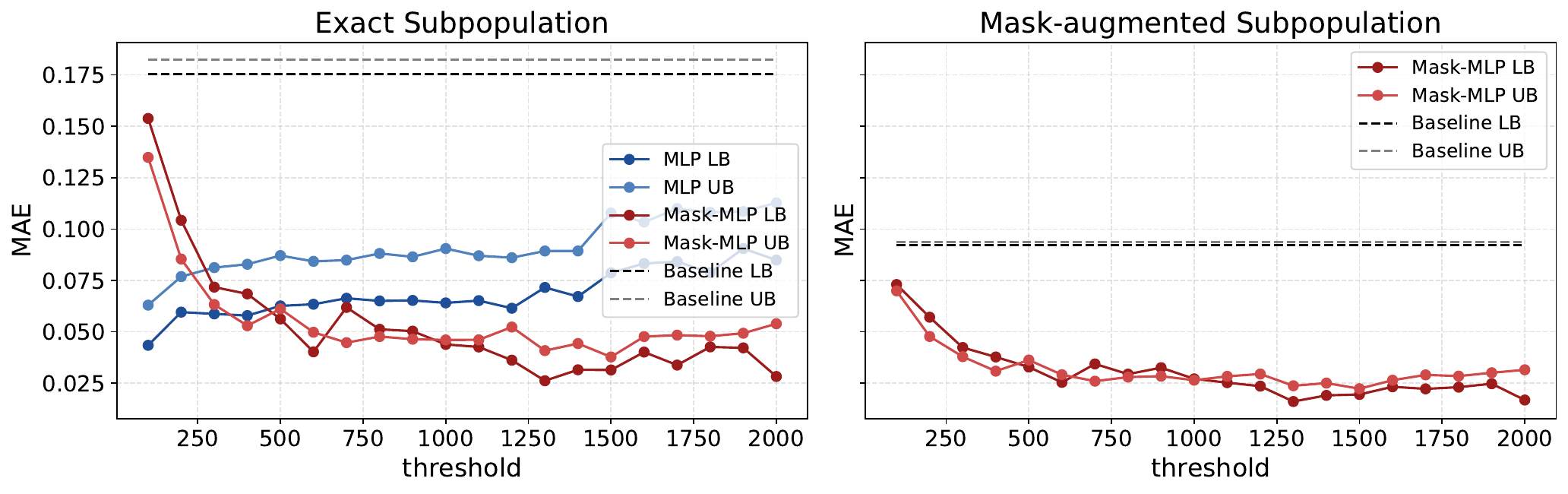}
            \subcaption{Covariate (200k)}
            \label{fig:MAE_cov_200k}
        \end{minipage}
        \\
        \vspace{-2mm} 
        \begin{minipage}[b]{0.5\textwidth}
            \centering
            \includegraphics[width=0.99\linewidth]{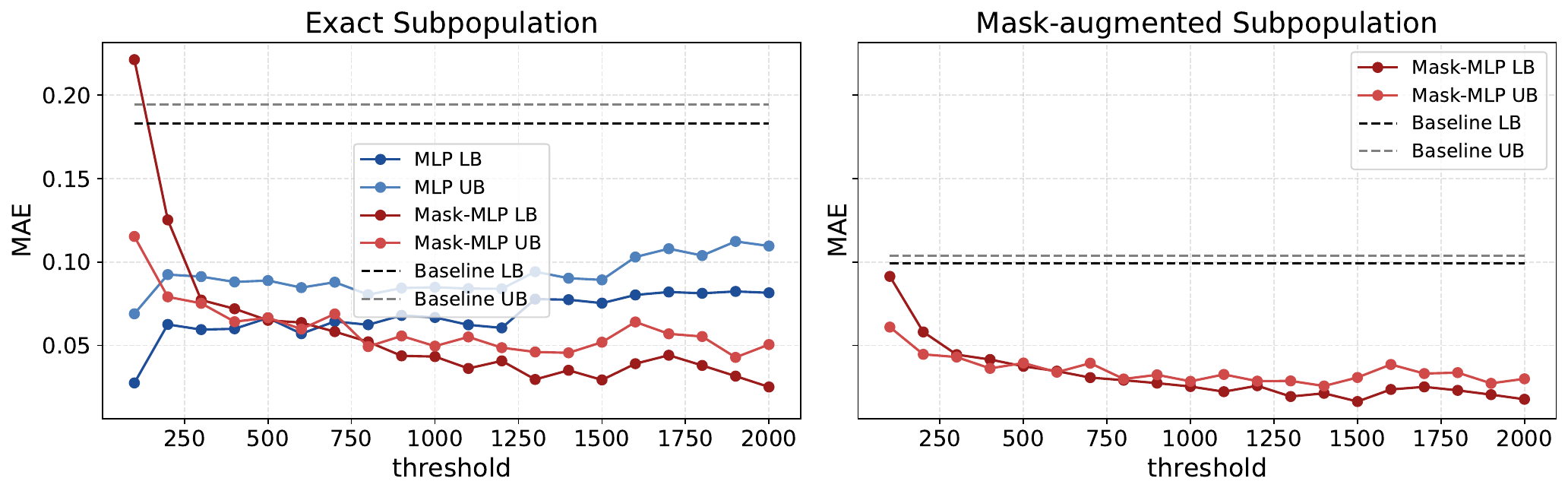}
            \subcaption{Confounder (200k)}
            \label{fig:MAE_conf_200k}
        \end{minipage}
        &
        \begin{minipage}[b]{0.5\textwidth}
            \centering
            \includegraphics[width=0.99\linewidth]{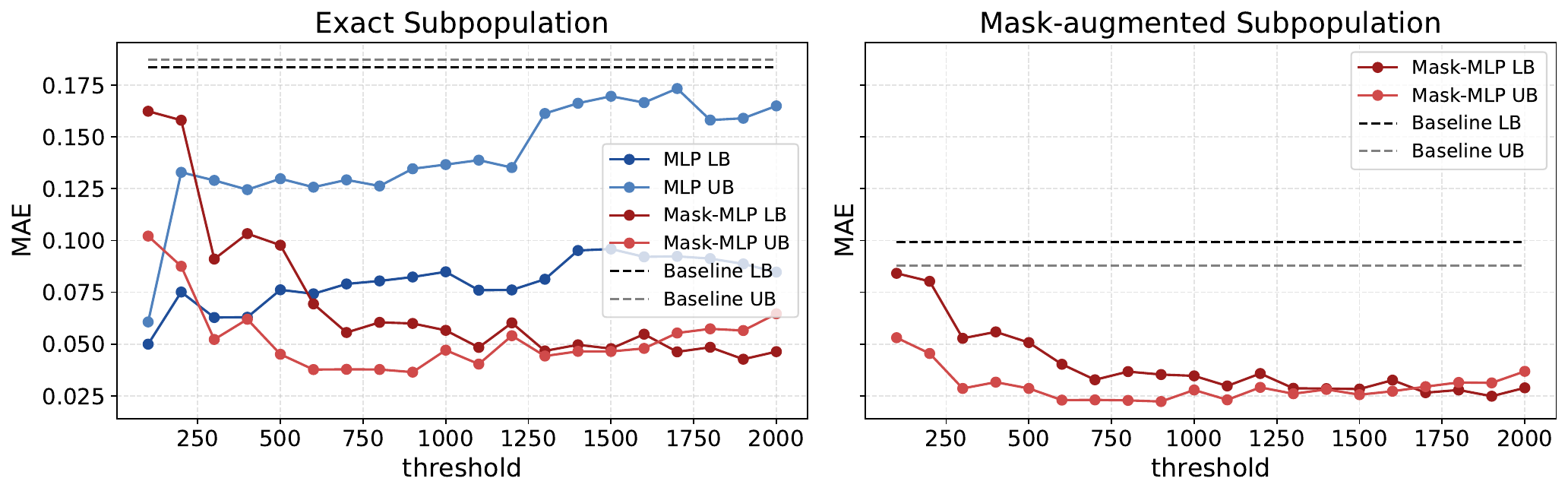}
            \subcaption{Mediator (200k)}
            \label{fig:MAE_med_200k}
        \end{minipage}
    \end{tabular}

    \caption{Threshold sweeps at the 200k budget across four SCMs. Each plot contains two panels: evaluation on exact queries (left) and on mask queries (right). All MAEs are mean values over five runs.}
    \label{fig:MAE_sweep_200k_all}
\end{figure*}

    We further analyze sensitivity to the threshold that filters low-support queries.
    Figure~\ref{fig:MAE_sweep_200k_all} shows representative threshold sweeps under the 200k budget (plots for other budgets are in Appendix~\ref{app:more_plots}). There is a trade-off across threshold settings: a lower threshold yields more training data but also cause more noise due to small sample size, while a higher threshold implies better data quality but generate less training data. Accordingly, across SCMs, Mask-MLP usually performs better at higher thresholds, whereas Exact-MLP often prefers lower thresholds. And we select best choice of threshold to represent the performance of the model. In all cases, both models outperform the baseline at their respective best thresholds.

    \subsection{Case Study: Treat-or-Not for a Toxic but Potentially Life-Saving Drug}
    \label{sec:drug-case}
    
    \paragraph{Background.}
    We consider a life-threatening disease for which a drug exists but has substantial toxicity. Each patient is described by ten observed binary covariates $Z_{1:10}$ (e.g., two bits for age $Z_1Z_2$, sex $Z_3$, and comorbidities $Z_4$–$Z_{10}$), while additional unobserved factors may also influence outcomes. We observe 200k experimental (RCT) records and 200k observational records at the population level. Now consider a new patient whose covariate profile $z^\star$ is fully specified. The patient wishes to know whether to take the drug. However, like 936 of the 1024 exact subpopulations, the hospital has fewer than 300 recorded patients in this subpopulation, which is too limited for reliably estimating PNS bounds using the plug-in baseline. Let $X$ denote treatment ($x$ = treated, $x'$ = untreated) and $Y$ denote the clinical outcome ($y$ = survived, $y'$ = deceased). Beyond the observed $Z_{1:10}$, a mediator $M$ (e.g., viral load) and further unobserved factors may also affect $Y$.
    
    \paragraph{Pipeline.}
    In this case, we have the number of observable covariates $d=10$. Therefore, we got exact queries $q\in\{0,1\}^d$ and mask-augmented subpopulations $q\in\{0,1,X\}^d$.
    Next, following Section~\ref{sec:method}, we identify well-sampled subpopulations among the $2^{10}$ exact groups and compute their reliable sample-based bounds using Eqs.~\eqref{pnslb}--\eqref{pnsub}.
    We then train two predictors:
    (i) an \textbf{MLP (Mish)} trained on threshold-filtered exact supervision,
    and (ii) a \textbf{Mask-MLP (Mish)} trained on threshold-filtered mask-augmented supervision over the full $3^{10}$ query space.
    Both predictors extrapolate \emph{feasible} PNS bounds to the long tail (including rare and unseen queries), outputting $\mathrm{PNS_{LB}}(z^\star)$ and $\mathrm{PNS_{UB}}(z^\star)$. At the same time, we also calculate PNS bounds via Eqs.~\eqref{pnslb}--\eqref{pnsub} based on his limited subpopulation.
    
    \paragraph{Decision rule.}
    We set a decision threshold $\theta=0.5$ (Treat iff $\mathrm{PNS}\ge 0.5$):
    we \emph{treat} when $\mathrm{LB}\ge\theta$,
    \emph{do not treat} when $\mathrm{UB}<\theta$,
    and mark \emph{uncertain} otherwise.
    We abbreviate decisions as \textbf{T} (treat), \textbf{N} (not treat), and \textbf{U} (unknown).
    
    \begin{table}[!t]
    \centering
    \caption{Decision at $\theta=0.5$ for the rare exact group $z=\texttt{1101001110}$.
    B: baseline (naive plug-in); M: MLP; MM: Mask-MLP.}
    \label{tab:drug_examples}
    \scriptsize
    \setlength{\tabcolsep}{1.3pt}
    \renewcommand{\arraystretch}{0.95}
    \begin{tabular}{@{}lccccc@{}}
    \toprule
     & \multicolumn{4}{c}{LB/UB} & \\
    \cmidrule(lr){2-5}
    $z$ (10-bit) & True & B & M & MM & Dec.\ (True/B/M/MM) \\
    \midrule
    \texttt{1101001110} & 0.223/0.275 & 1.000/1.000 & 0.353/0.591 & 0.199/0.289 & \textbf{N}/T/U/\textbf{N} \\
    \bottomrule
    \end{tabular}
    \end{table}

    \begin{table}[!t]
    \centering
    \caption{Single-missing variants of $z=\texttt{1101001110}$ under $\theta=0.5$.
    Each row replaces one coordinate by $X$ (unknown).
    We report oracle (True), baseline (B), and Mask-MLP (MM) LB/UB, plus decisions (T/N/U).}
    \label{tab:drug_missing_one}
    \scriptsize
    \setlength{\tabcolsep}{3.5pt}   
    \renewcommand{\arraystretch}{1.05} 
    \begin{tabular}{@{}l c c c c@{}}
    \toprule
     & \multicolumn{3}{c}{LB/UB} & \\
    \cmidrule(lr){2-4}
    Masked query & True & B & MM & Dec.\ (True/B/MM) \\
    \midrule
    \texttt{X101001110}  & 0.164/0.199 & 1.000/1.000 & 0.161/0.225 & \textbf{N}/T/\textbf{N} \\
    \texttt{1X01001110}  & 0.278/0.435 & 0.333/0.333 & 0.268/0.474 & \textbf{N}/\textbf{N}/\textbf{N} \\
    \texttt{11X1001110}  & 0.140/0.165 & 0.667/0.667 & 0.110/0.193 & \textbf{N}/T/\textbf{N} \\
    \texttt{110X001110}  & 0.308/0.491 & 0.361/0.444 & 0.294/0.451 & \textbf{N}/\textbf{N}/\textbf{N} \\
    \texttt{1101X01110}  & 0.324/0.528 & 0.324/0.532 & 0.382/0.528 & \textbf{U}/\textbf{U}/\textbf{U} \\
    \texttt{11010X1110}  & 0.170/0.205 & 0.000/0.000 & 0.170/0.227 & \textbf{N}/\textbf{N}/\textbf{N} \\
    \texttt{110100X110}  & 0.225/0.278 & 1.000/1.000 & 0.196/0.292 & \textbf{N}/T/\textbf{N} \\
    \texttt{1101001X10}  & 0.250/0.343 & 1.000/1.000 & 0.218/0.337 & \textbf{N}/T/\textbf{N} \\
    \texttt{11010011X0}  & 0.260/0.384 & 0.400/0.400 & 0.226/0.418 & \textbf{N}/\textbf{N}/\textbf{N} \\
    \texttt{110100111X}  & 0.222/0.273 & 1.000/1.000 & 0.196/0.286 & \textbf{N}/T/\textbf{N} \\
    \bottomrule
    \end{tabular}
    \end{table}

    \paragraph{Illustrative example under $\theta=0.5$.}
    Consider the patient with $z=\texttt{1101001110}$ in Table~\ref{tab:drug_examples}.
    According to the oracle (true) bounds ($\mathrm{LB}=0.2229$ and $\mathrm{UB}=0.2746$), we have $\mathrm{UB}<0.5$, which means the patient should \textbf{not be treated}.
    In contrast, the baseline plug-in method outputs $\mathrm{LB}=\mathrm{UB}=1.0000$ under extreme data sparsity, and suggests \textbf{treatment} with full confidence.
    As for the MLP, it gives an uncertain conclusion, since its interval crosses the decision threshold ($\mathrm{LB}=0.3529<0.5<\mathrm{UB}=0.5908$).
    Finally, Mask-MLP predicts $\mathrm{LB}=0.1994$ and $\mathrm{UB}=0.2890$, and correctly recommend \textbf{no treatment}, consistent with the oracle.
    
    \paragraph{Robustness under a single missing feature.}
    In practice, a patient may have missing or unclear feature (may caused by privacy issue or just unsure). We therefore construct ten \emph{single-missing} variants by replacing exactly one coordinate of $z=\texttt{1101001110}$ with $X$ (unknown).
    Table~\ref{tab:drug_missing_one} reports the oracle bounds, the baseline predictions, and the Mask-MLP predictions.
    Among the ten single-missing variants, nine cases have the same output with exact subpopulation prediction, \emph{only one} case becomes \textbf{unknown}. However, it is not caused by wrong predication: the oracle also has the same conclusion.
    This is the desired pattern: robustness to minor missingness, together with uncertainty flags if the query is intrinsically ambiguous.

\section{Discussion}\label{sec:discuss}
    We studied whether our machine learning models can predict \emph{oracle} PNS bounds for subpopulations with insufficient data based on finite experimental and observational samples.
    Across four SCMs and multiple sample budgets and the study case, we can draw such conclusion.: (i) learning-based predictors dramatically improve over naive plug-in estimation under limited data, and (ii) Mask-MLP typically strengthens generalization and further supports query space.

    \paragraph{Model comparison and underlying reasons.}
    Our most important issue is data size, or, more specifically, the sample budget. In the ideal scenario, assume that unlimited data can be provided; we can calculate PNS bounds via formulas based on oracle experimental and conditional probabilities. So, the baseline method is the best with sufficient data. If we consider the cost of data, the machine learning models could be better than the baseline. And the situation is also similar for the difference between Exact-MLP and Mask-MLP on exact subpopulation predictions. When the sample budget is relatively small for the SCM, Mask-MLP can be trained with much more mask-augmented data, which performs better. But when the data size is relatively large for the SCM, the MLP can learn from enough training data and is able to predict more accurately. In our case, when the sample budget is lower than 500k, Mask-MLP is more accurate and stable (except for the Direct SCM, which is too simple).

    \paragraph{Decision making.}
    In addition, our case study shows how this model can be applied in practice. In the case study, the conclusion computed directly using the Tian--Pearl formulas is completely opposite to the ground truth; Exact-MLP is uncertain, whereas Mask-MLP remains consistent with the ground truth. Moreover, when a single covariate is missing at decision time, only one of the ten single-missing variants is intrinsically ambiguous under the oracle (and this is also accurately identified by our model). This pattern is ideal for high-stakes settings: it is robust to minor missingness and raises uncertainty flags only when the underlying query itself has inherent ambiguity.

    \paragraph{From full coverage to targeted decisions.}
    A fundamental goal in causal decision making is to identify subpopulations with sufficiently large $\mathrm{PNS}_{\mathrm{LB}}$ (or small $\mathrm{PNS}_{\mathrm{UB}}$) rather than to perfectly predict every subpopulation. Hence, accurately predicting all subpopulations is unnecessary for practical use. An interesting direction is to determine a minimal training set that reliably recovers the top-$k$ subpopulations under budget or risk constraints.
    
    \paragraph{Limitations and future directions.}
    Though our experiment is enough to show the availability of predicting PNS bounds via machine learning, we still leave a lot of space for more experiments. The first is data size for more complex SCMs: a 200k sample budget and Mask-MLP with Mish are enough to render good results in our SCM. But when the SCM is more complex, or more covariates are included, the sample budget may need to be changed. The second is the choice of ML models: although MLP is the best in our past experiment under our settings, for more complex SCMs, richer covariates, and some more complex model (like Transformer) may perform better than MLP. So, exploring more complex SCMs is important in the future.
    
    An additional limitation is that on real-world data, oracle test bounds are generally unavailable, making it difficult to directly validate bound tightness and calibration beyond indirect or downstream evaluations. A potential way is to train and test only on reliable subpopulations.

\section{Conclusion}\label{sec:conclusion}

    In this paper, we show that PNS bounds can be learned and extrapolated to subpopulations with insufficient support from finite experimental and observational data. We propose two predictors, Exact-MLP and Mask-MLP, and empirically demonstrate that both consistently outperform naive plug-in estimation based on the Tian–Pearl bounds across four SCMs, on both exact subpopulation and mask queries. In low-data regimes, Mask-MLP typically achieves better accuracy than Exact-MLP, reflecting the benefit of mask-augmented supervision. This is also reflected in our examples: Mask-MLP showed great potential in real-world decision making based on PNS.

    In the future, we can explore more complex SCMs and investigate whether more complex models would perform better. Furthermore, overcoming the lack of oracle data in real-world datasets is also a promising direction.

\nocite{langley00}

\bibliography{example_paper}
\bibliographystyle{icml2026}

\newpage
\onecolumn
\appendix
\section{Appendix / supplemental material}
\subsection{The Causal Model}
    There are four SCMs mentioned in our work, which are Confounder, Covariate, Direct and Mediator. And their SCM are shown in figure \ref{apfig:different_scm}.
    \begin{figure}[h]
        \centering
        \begin{subfigure}[b]{0.2\textwidth}
            \centering
            \begin{tikzpicture}[->,>=stealth',node distance=2cm,
              thick,main node/.style={circle,fill,inner sep=1.5pt}]
              \node[main node] (1) [label=above:{$Z$}]{};
              \node[main node] (2) [below left =1cm of 1,label=left:$X$]{};
              \node[main node] (3) [below right =1cm of 1,label=right:$Y$] {};
              \path[every node/.style={font=\sffamily\small}]
                (1) edge node {} (2)
                (1) edge node {} (3)
                (2) edge node {} (3);
            \end{tikzpicture}
            \caption{Confounder $Z$ with direct effect}
            \label{apfig:confounder}
        \end{subfigure}
        \hfill
        \begin{subfigure}[b]{0.2\textwidth}
            \centering
            \begin{tikzpicture}[->,>=stealth',node distance=2cm,
              thick,main node/.style={circle,fill,inner sep=1.5pt}]
              \node[main node] (1) [label=above:{$Z$}]{};
              \node[main node] (2) [below left =1cm of 1,label=left:$X$]{};
              \node[main node] (3) [below right =1cm of 1,label=right:$Y$] {};
              \path[every node/.style={font=\sffamily\small}]
                (1) edge node {} (3)
                (2) edge node {} (3);
            \end{tikzpicture}
            \caption{Covariate $Z$}
            \label{apfig:outcome covariate}
        \end{subfigure}
        \hfill
        \begin{subfigure}[b]{0.2\textwidth}
            \centering
            \begin{tikzpicture}[->,>=stealth',node distance=2cm,
              thick,main node/.style={circle,fill,inner sep=1.5pt}]
              \node[main node] (1) [label=above:{$Z$}]{};
              \node[main node] (2) [below left =1cm of 1,label=left:$X$]{};
              \node[main node] (3) [below right =1cm of 1,label=right:$Y$] {};
              \path[every node/.style={font=\sffamily\small}]
                (2) edge node {} (3);
            \end{tikzpicture}
            \caption{Direct}
            \label{apfig:direct effect}
        \end{subfigure}
        \hfill
        \begin{subfigure}[b]{0.2\textwidth}
            \centering
            \begin{tikzpicture}[->,>=stealth',node distance=1.4cm,
              thick,main node/.style={circle,fill,inner sep=1.5pt}]
              \node[main node] (1) [label=above:{$Z$}]{};
              \node[main node] (2) [below left =0.4cm and 0.7cm of 1,label=left:$X$]{};
              \node[main node] (3) [below right =0.4cm and 0.7cm of 1,label=right:$Y$]{};
              \node[main node] (4) [below =0.8cm of 1,label=below:$M$]{};
              \path[every node/.style={font=\sffamily\small}]
                (1) edge (2)
                (1) edge (3)
                (2) edge (3)
                (2) edge (4)
                (4) edge (3);
            \end{tikzpicture}
            \caption{Mediator $M$ and direct effect}
            \label{apfig:mediator_with_direct}
        \end{subfigure}
        \caption{Different SCMs in this study.}
        \label{apfig:different_scm}
    \end{figure}

\subsubsection{Confounder}

    \textbf{Confounder} SCM is shown in figure \ref{apfig:confounder}.
    The coefficients for \( X_Z, Y_Z \), and \( C_Y \) were uniformly generated from the range \([-1,1]\), while the parameters of the Bernoulli distribution were uniformly generated from \([0,1]\). The detailed model is as follows:
    \begin{eqnarray*}
        &&\begin{cases}
            Z_i &= U_{Z_i} \text{ for } i \in \{1,...,20\},\\
            X&=f_X(X_Z,U_X)\\
            &=\begin{cases}
                1& \text{ if } X_Z+U_X > 0.5\\
                0& \text{ otherwise, }\\
            \end{cases}\\
            Y&=f_Y(X,Y_Z,U_Y)\\
            &=\begin{cases}
                1& \text{ if } 0<C_Y \cdot X+Y_Z+U_Y <1 \\
                1& \text{ if } 1<C_Y \cdot X+Y_Z+U_Y <2 \\
                0& \text{ otherwise. }\\
            \end{cases}
        \end{cases}\\
    &&\text{where, } U_{Z_i}, U_X, U_Y \text{ are binary exogenous variables with Bernoulli distributions.}\\
    &&s.t., \\
    &X_Z& =
    \begin{bmatrix}
    Z_1~Z_2~...~Z_{20}
    \end{bmatrix}\times
    \begin{bmatrix}
    0.259223510143\\ -0.658140989167\\ -0.75025831768\\ 0.162906462426\\ 0.652023463285\\ -0.0892939586541\\ 0.421469107769\\ -0.443129684766\\ 0.802624388789\\ -0.225740978499\\ 0.716621631717\\ 0.0650682260309\\ -0.220690334026\\ 0.156355773665\\ -0.50693672491\\ -0.707060278115\\ 0.418812816935\\ -0.0822118703986\\ 0.769299853833\\ -0.511585391002
    \end{bmatrix},
    Y_Z =
    \begin{bmatrix}
    Z_1~Z_2~...~Z_{20}
    \end{bmatrix}\times
    \begin{bmatrix}
    -0.792867111918\\ 0.759967136147\\ 0.55437722369\\ 0.503970540409\\ -0.527187144651\\ 0.378619988091\\ 0.269255196301\\ 0.671597043594\\ 0.396010142274\\ 0.325228576643\\ 0.657808327574\\ 0.801655023993\\ 0.0907679484097\\ -0.0713852594543\\ -0.0691046005285\\ -0.222582013343\\ -0.848408031595\\ -0.584285069026\\ -0.324874831799\\ 0.625621583197
    \end{bmatrix}
    \end{eqnarray*}
    \begin{eqnarray*}
    &&U_{Z_1} \sim \text{Bernoulli}(0.352913861526), U_{Z_2} \sim \text{Bernoulli}(0.460995855543),\\
    &&U_{Z_3} \sim \text{Bernoulli}(0.331702473392), U_{Z_4} \sim \text{Bernoulli}(0.885505026779),\\
    &&U_{Z_5} \sim \text{Bernoulli}(0.017026872706), U_{Z_6} \sim \text{Bernoulli}(0.380772701708),\\
    &&U_{Z_7} \sim \text{Bernoulli}(0.028092602705), U_{Z_8} \sim \text{Bernoulli}(0.220819399962),\\
    &&U_{Z_9} \sim \text{Bernoulli}(0.617742227477), U_{Z_{10}} \sim \text{Bernoulli}(0.981975046713),\\
    &&U_{Z_{11}} \sim \text{Bernoulli}(0.142042291381), U_{Z_{12}} \sim \text{Bernoulli}(0.833602592350),\\
    &&U_{Z_{13}} \sim \text{Bernoulli}(0.882938907115), U_{Z_{14}} \sim \text{Bernoulli}(0.542143191999),\\
    &&U_{Z_{15}} \sim \text{Bernoulli}(0.085023436884), U_{Z_{16}} \sim \text{Bernoulli}(0.645357252864),\\
    &&U_{Z_{17}} \sim \text{Bernoulli}(0.863787135134), U_{Z_{18}} \sim \text{Bernoulli}(0.460539711624),\\
    &&U_{Z_{19}} \sim \text{Bernoulli}(0.314014079207), U_{Z_{20}} \sim \text{Bernoulli}(0.685879388218),\\
    &&U_{X} \sim \text{Bernoulli}(0.601680857267), U_{Y} \sim \text{Bernoulli}(0.497668975278),\\
    &&C_Y=-0.77953605542.
    \end{eqnarray*}

\subsubsection{Covariate}
    \textbf{Covariate} SCM is shown in figure \ref{apfig:outcome covariate}.
    The coefficients for \(Y_Z \) and \( C_Y \) were uniformly generated from the range \([-1,1]\), while the parameters of the Bernoulli distribution were uniformly generated from \([0,1]\). The detailed model is as follows:
    \begin{eqnarray*}
        &&\begin{cases}
            Z_i &= U_{Z_i} \text{ for } i \in \{1,...,20\},\\
            X&=f_X(U_X)\\
            &=\begin{cases}
                1& \text{ if } U_X > 0.5\\
                0& \text{ otherwise, }\\
            \end{cases}\\
            Y&=f_Y(X,Y_Z,U_Y)\\
            &=\begin{cases}
                1& \text{ if } 0<C_Y \cdot X+Y_Z+U_Y <1 \\
                1& \text{ if } 1<C_Y \cdot X+Y_Z+U_Y <2 \\
                0& \text{ otherwise. }\\
            \end{cases}
        \end{cases}\\
    &&\text{where, } U_{Z_i}, U_X, U_Y \text{ are binary exogenous variables with Bernoulli distributions.}\\
    &&s.t., \\
    &Y_Z& =
    \begin{bmatrix}
    Z_1~Z_2~...~Z_{20}
    \end{bmatrix}\times
    \begin{bmatrix}
    -0.792867111918\\ 0.759967136147\\ 0.55437722369\\ 0.503970540409\\ -0.527187144651\\ 0.378619988091\\ 0.269255196301\\ 0.671597043594\\ 0.396010142274\\ 0.325228576643\\ 0.657808327574\\ 0.801655023993\\ 0.0907679484097\\ -0.0713852594543\\ -0.0691046005285\\ -0.222582013343\\ -0.848408031595\\ -0.584285069026\\ -0.324874831799\\ 0.625621583197
    \end{bmatrix}
    \end{eqnarray*}
    \begin{eqnarray*}
    &&U_{Z_1} \sim \text{Bernoulli}(0.352913861526), U_{Z_2} \sim \text{Bernoulli}(0.460995855543),\\
    &&U_{Z_3} \sim \text{Bernoulli}(0.331702473392), U_{Z_4} \sim \text{Bernoulli}(0.885505026779),\\
    &&U_{Z_5} \sim \text{Bernoulli}(0.017026872706), U_{Z_6} \sim \text{Bernoulli}(0.380772701708),\\
    &&U_{Z_7} \sim \text{Bernoulli}(0.028092602705), U_{Z_8} \sim \text{Bernoulli}(0.220819399962),\\
    &&U_{Z_9} \sim \text{Bernoulli}(0.617742227477), U_{Z_{10}} \sim \text{Bernoulli}(0.981975046713),\\
    &&U_{Z_{11}} \sim \text{Bernoulli}(0.142042291381), U_{Z_{12}} \sim \text{Bernoulli}(0.833602592350),\\
    &&U_{Z_{13}} \sim \text{Bernoulli}(0.882938907115), U_{Z_{14}} \sim \text{Bernoulli}(0.542143191999),\\
    &&U_{Z_{15}} \sim \text{Bernoulli}(0.085023436884), U_{Z_{16}} \sim \text{Bernoulli}(0.645357252864),\\
    &&U_{Z_{17}} \sim \text{Bernoulli}(0.863787135134), U_{Z_{18}} \sim \text{Bernoulli}(0.460539711624),\\
    &&U_{Z_{19}} \sim \text{Bernoulli}(0.314014079207), U_{Z_{20}} \sim \text{Bernoulli}(0.685879388218),\\
    &&U_{X} \sim \text{Bernoulli}(0.601680857267), U_{Y} \sim \text{Bernoulli}(0.497668975278),\\
    &&C_Y=-0.77953605542.
    \end{eqnarray*}
\subsubsection{Direct}
    \textbf{Direct} SCM is shown in figure \ref{apfig:direct effect}.
    The coefficient for \( C_Y \) were uniformly generated from the range \([-1,1]\), while the parameters of the Bernoulli distribution were uniformly generated from \([0,1]\). The detailed model is as follows:
    \begin{eqnarray*}
        &&\begin{cases}
            Z_i &= U_{Z_i} \text{ for } i \in \{1,...,20\},\\
            X&=f_X(U_X)\\
            &=\begin{cases}
                1& \text{ if } U_X > 0.5\\
                0& \text{ otherwise, }\\
            \end{cases}\\
            Y&=f_Y(X,U_Y)\\
            &=\begin{cases}
                1& \text{ if } C_Y \cdot X+U_Y > 0.7 \\
                0& \text{ otherwise. }\\
            \end{cases}
        \end{cases}\\
    &&\text{where, } U_{Z_i}, U_X, U_Y \text{ are binary exogenous variables with Bernoulli distributions.}\\
    \end{eqnarray*}
    \begin{eqnarray*}
    &&U_{Z_1} \sim \text{Bernoulli}(0.352913861526), U_{Z_2} \sim \text{Bernoulli}(0.460995855543),\\
    &&U_{Z_3} \sim \text{Bernoulli}(0.331702473392), U_{Z_4} \sim \text{Bernoulli}(0.885505026779),\\
    &&U_{Z_5} \sim \text{Bernoulli}(0.017026872706), U_{Z_6} \sim \text{Bernoulli}(0.380772701708),\\
    &&U_{Z_7} \sim \text{Bernoulli}(0.028092602705), U_{Z_8} \sim \text{Bernoulli}(0.220819399962),\\
    &&U_{Z_9} \sim \text{Bernoulli}(0.617742227477), U_{Z_{10}} \sim \text{Bernoulli}(0.981975046713),\\
    &&U_{Z_{11}} \sim \text{Bernoulli}(0.142042291381), U_{Z_{12}} \sim \text{Bernoulli}(0.833602592350),\\
    &&U_{Z_{13}} \sim \text{Bernoulli}(0.882938907115), U_{Z_{14}} \sim \text{Bernoulli}(0.542143191999),\\
    &&U_{Z_{15}} \sim \text{Bernoulli}(0.085023436884), U_{Z_{16}} \sim \text{Bernoulli}(0.645357252864),\\
    &&U_{Z_{17}} \sim \text{Bernoulli}(0.863787135134), U_{Z_{18}} \sim \text{Bernoulli}(0.460539711624),\\
    &&U_{Z_{19}} \sim \text{Bernoulli}(0.314014079207), U_{Z_{20}} \sim \text{Bernoulli}(0.685879388218),\\
    &&U_{X} \sim \text{Bernoulli}(0.601680857267), U_{Y} \sim \text{Bernoulli}(0.497668975278),\\
    &&C_Y=-0.77953605542.
    \end{eqnarray*}
\subsubsection{Mediator}
    \textbf{Mediator} SCM is shown in figure \ref{apfig:confounder}.
    The coefficients for \( X_Z, Y_Z, C_M, C_{Y_M} \), and \( C_{Y_X} \) were uniformly generated from the range \([-1,1]\), while the parameters of the Bernoulli distribution were uniformly generated from \([0,1]\). The detailed model is as follows:
    \begin{eqnarray*}
        &&\begin{cases}
            Z_i &= U_{Z_i} \text{ for } i \in \{1,...,20\},\\
            X&=f_X(X_Z,U_X)\\
            &=\begin{cases}
                1& \text{ if } X_Z+U_X > 0.5\\
                0& \text{ otherwise, }\\
            \end{cases}\\

            M&=f_M(X,U_M)\\
            &=\begin{cases}
                1& \text{ if } C_M \cdot X + U_M > 0.5\\
                0& \text{ otherwise, }\\
            \end{cases}\\
            
            Y&=f_Y(X, M, Y_Z, U_Y)\\
            &=\begin{cases}
                1& \text{ if } C_{Y_X} \cdot X + C_{Y_M} \cdot M+Y_Z+U_Y > 2 \\
                0& \text{ otherwise. }\\
            \end{cases}
        \end{cases}\\
    &&\text{where, } U_{Z_i}, U_X, U_Y ,U_M \text{ are binary exogenous variables with Bernoulli distributions.}\\
    &&s.t., \\
    &X_Z& =
    \begin{bmatrix}
    Z_1~Z_2~...~Z_{20}
    \end{bmatrix}\times
    \begin{bmatrix}
    0.259223510143\\ -0.658140989167\\ -0.75025831768\\ 0.162906462426\\ 0.652023463285\\ -0.0892939586541\\ 0.421469107769\\ -0.443129684766\\ 0.802624388789\\ -0.225740978499\\ 0.716621631717\\ 0.0650682260309\\ -0.220690334026\\ 0.156355773665\\ -0.50693672491\\ -0.707060278115\\ 0.418812816935\\ -0.0822118703986\\ 0.769299853833\\ -0.511585391002
    \end{bmatrix},
    Y_Z =
    \begin{bmatrix}
    Z_1~Z_2~...~Z_{20}
    \end{bmatrix}\times
    \begin{bmatrix}
    -0.792867111918\\ 0.759967136147\\ 0.55437722369\\ 0.503970540409\\ -0.527187144651\\ 0.378619988091\\ 0.269255196301\\ 0.671597043594\\ 0.396010142274\\ 0.325228576643\\ 0.657808327574\\ 0.801655023993\\ 0.0907679484097\\ -0.0713852594543\\ -0.0691046005285\\ -0.222582013343\\ -0.848408031595\\ -0.584285069026\\ -0.324874831799\\ 0.625621583197
    \end{bmatrix}
    \end{eqnarray*}
    \begin{eqnarray*}
    &&U_{Z_1} \sim \text{Bernoulli}(0.352913861526), U_{Z_2} \sim \text{Bernoulli}(0.460995855543),\\
    &&U_{Z_3} \sim \text{Bernoulli}(0.331702473392), U_{Z_4} \sim \text{Bernoulli}(0.885505026779),\\
    &&U_{Z_5} \sim \text{Bernoulli}(0.017026872706), U_{Z_6} \sim \text{Bernoulli}(0.380772701708),\\
    &&U_{Z_7} \sim \text{Bernoulli}(0.028092602705), U_{Z_8} \sim \text{Bernoulli}(0.220819399962),\\
    &&U_{Z_9} \sim \text{Bernoulli}(0.617742227477), U_{Z_{10}} \sim \text{Bernoulli}(0.981975046713),\\
    &&U_{Z_{11}} \sim \text{Bernoulli}(0.142042291381), U_{Z_{12}} \sim \text{Bernoulli}(0.833602592350),\\
    &&U_{Z_{13}} \sim \text{Bernoulli}(0.882938907115), U_{Z_{14}} \sim \text{Bernoulli}(0.542143191999),\\
    &&U_{Z_{15}} \sim \text{Bernoulli}(0.085023436884), U_{Z_{16}} \sim \text{Bernoulli}(0.645357252864),\\
    &&U_{Z_{17}} \sim \text{Bernoulli}(0.863787135134), U_{Z_{18}} \sim \text{Bernoulli}(0.460539711624),\\
    &&U_{Z_{19}} \sim \text{Bernoulli}(0.314014079207), U_{Z_{20}} \sim \text{Bernoulli}(0.685879388218),\\
    &&U_{X} \sim \text{Bernoulli}(0.698319142733), U_{Y} \sim \text{Bernoulli}(0.502331024722),\\
    &&U_{M} \sim \text{Bernoulli}(0.402331024722), C_M = -0.74234511918.\\
    &&C_{Y_M} = 0.24235642321, C_{Y_X} = 0.87953605542.
    \end{eqnarray*}
\subsection{Detailed Data Generating Process}

\subsubsection{Confounder}
    \textbf{Confounder} SCM (Figure~\ref{apfig:confounder}).  
    When all 20 binary features are observed, let \( z=(z_1,\dots,z_{20}) \) and denote
    \( X_Z(z)\) and \(Y_Z(z)\) as fixed constants.  The causal quantities are
    
    \begin{eqnarray*}
    PNS(z) &=& P\!\bigl(Y=0_{X=0},\,Y=1_{X=1}\mid z\bigr)  \\
           &=& P(U_Y=0)\,T_0 + P(U_Y=1)\,T_1, \\[4pt]
    T_0 &=&
    \begin{cases}
    1,& \text{if } f_Y(0,Y_Z(z),0)=0 \text{ and } f_Y(1,Y_Z(z),0)=1,\\
    0,& \text{otherwise},
    \end{cases}\\[4pt]
    T_1 &=&
    \begin{cases}
    1,& \text{if } f_Y(0,Y_Z(z),1)=0 \text{ and } f_Y(1,Y_Z(z),1)=1,\\
    0,& \text{otherwise}.
    \end{cases}
    \end{eqnarray*}
    
    \begin{eqnarray*}
    P\!\bigl(Y=1\mid do(X),z\bigr)
      &=& \sum_{u_y} P(U_Y=u_y)\,f_Y\!\bigl(X,Y_Z(z),u_y\bigr),\\
    P\!\bigl(Y=1\mid X,z\bigr)
      &=& \sum_{u_x,u_y} P(U_X=u_x)P(U_Y=u_y)\,
          f_Y\!\bigl(f_X(X_Z(z),u_x),Y_Z(z),u_y\bigr).
    \end{eqnarray*}
    
    If only the first 10 features are observed, let  
    \(c=(z_1,\dots,z_{10})\).  Each subpopulation \(c\) contains
    \(2^{10}=1024\) full feature vectors \(s_0,\dots,s_{1023}\).  Then
    
    \begin{eqnarray*}
    PNS_{\text{subpop}}(c) &=& \sum_{j=0}^{1023} \frac{P(s_j)}{P(c)}\,PNS(s_j),\\
    P\!\bigl(Y=1\mid do(X),c\bigr) &=& \sum_{j=0}^{1023} \frac{P(s_j)}{P(c)}\,
                                         P\!\bigl(Y=1\mid do(X),s_j\bigr),\\
    P\!\bigl(Y=1\mid X,c\bigr)     &=& \sum_{j=0}^{1023} \frac{P(s_j)}{P(c)}\,
                                         P\!\bigl(Y=1\mid X,s_j\bigr),
    \end{eqnarray*}
    where \(P(s_j)\) is the product of the Bernoulli priors for \(Z_{11{:}20}\).
    
\subsubsection{Covariate}
    \textbf{Outcome‑Covariate} SCM (Figure~\ref{apfig:outcome covariate}).  
    Here \(X\) is independent of \(Z\); there is no \(X_Z\).  For any
    \(z\):
    
    \begin{eqnarray*}
    PNS(z) &=& P(U_Y=0)\,T_0 + P(U_Y=1)\,T_1, \quad(\text{definitions as above}),\\
    P\!\bigl(Y=1\mid do(X),z\bigr)
           &=& \sum_{u_y} P(U_Y=u_y)\,f_Y\!\bigl(X,Y_Z(z),u_y\bigr),\\
    P\!\bigl(Y=1\mid X,z\bigr)
           &=& \sum_{u_x,u_y} P(U_X=u_x)P(U_Y=u_y)\,
               f_Y\!\bigl(f_X(u_x),Y_Z(z),u_y\bigr).
    \end{eqnarray*}
    
    Subpopulation‑level formulas are identical to those in the Confounder case.
    
\subsubsection{Direct}
    \textbf{Direct} SCM (Figure~\ref{apfig:direct effect}).  
    Here \(Z\) influences neither \(X\) nor \(Y\); there is no \(Y_Z\):
    
    \begin{eqnarray*}
    PNS(z) &=& P(U_Y=0)\,T_0 + P(U_Y=1)\,T_1,\\
    T_i &=&
    \begin{cases}
    1,& \text{if } f_Y(0,i)=0 \text{ and } f_Y(1,i)=1,\\
    0,& \text{otherwise},
    \end{cases}\\
    P\!\bigl(Y=1\mid do(X),z\bigr)
           &=& \sum_{u_y} P(U_Y=u_y)\,f_Y(X,u_y),\\
    P\!\bigl(Y=1\mid X,z\bigr)
           &=& \sum_{u_x,u_y} P(U_X=u_x)P(U_Y=u_y)\,
                f_Y\!\bigl(f_X(u_x),u_y\bigr).
    \end{eqnarray*}
    
    Because \(Z\) is irrelevant, the weighted‑average step over
    \(s_0{:}s_{1023}\) is omitted.
    
\subsubsection{Mediator}
    \textbf{Mediator} SCM (Figure~\ref{apfig:mediator_with_direct}).  
    For any full feature vector \(z\):
    
    \begin{eqnarray*}
    X &=& f_X\!\bigl(X_Z(z),U_X\bigr),\\
    M &=& f_M(X,U_M),\\
    Y &=& f_Y\!\bigl(X,M,Y_Z(z),U_Y\bigr).
    \end{eqnarray*}
    
    \paragraph{Individual‑level quantities}
    \begin{eqnarray*}
    PNS(z) &=& \sum_{u_x,u_m,u_y} P(U_X)P(U_M)P(U_Y)\,
            \mathbf{1}\!\bigl[
                f_Y(0,f_M(0,u_m),Y_Z(z),u_y)=0 \land \\
    &&\hspace{3.1cm}
                f_Y(1,f_M(1,u_m),Y_Z(z),u_y)=1
            \bigr],\\
    P\!\bigl(Y=1\mid do(X),z\bigr)
           &=& \sum_{u_m,u_y} P(U_M)P(U_Y)\,
               f_Y\!\bigl(X,f_M(X,u_m),Y_Z(z),u_y\bigr),\\
    P\!\bigl(Y=1\mid X,z\bigr)
           &=& \sum_{u_x,u_m,u_y} P(U_X)P(U_M)P(U_Y)\,
               f_Y\!\bigl(f_X(X_Z(z),u_x),\,f_M(f_X(X_Z(z),u_x),u_m),\\
    &&\hspace{5.5cm} Y_Z(z),u_y\bigr).
    \end{eqnarray*}
    
    \paragraph{Subpopulation‑level quantities}
    With only \(Z_1{:}Z_{10}\) observed, define
    \(c=(z_1,\dots,z_{10})\) and enumerate
    \(s_0{:}s_{1023}\) as before.  The three weighted‑average formulas are
    identical to those in the Confounder subsection, with the
    individual‑level expressions above substituted for
    \(PNS(s_j)\), \(P(Y=1\mid do(X),s_j)\), and
    \(P(Y=1\mid X,s_j)\).

\subsubsection{Mask-augmented query construction}
\label{app:mask_query}

We extend exact subpopulation queries $q\in\{0,1\}^{d}$ to mask-augmented queries
$q\in\{0,1,X\}^{d}$, where $q_j=X$ indicates that the $j$-th covariate is unspecified
(i.e., marginalized) in the query.
Let $\mathcal{Z}(q)\subseteq\{0,1\}^{d}$ denote the set of exact assignments consistent with $q$
(obtained by filling each $X$ position with $\{0,1\}$).

\paragraph{Oracle aggregation (SCM).}
For any oracle quantity $g(z)$ defined on exact assignments (e.g., $P(y\mid do(x),z)$ or
the oracle bound endpoints computed from SCM semantics), we define its value for a mask query
$q$ by mixture aggregation over consistent exact subpopulations:
\begin{equation}
\pi(q) \;=\; \sum_{z\in\mathcal{Z}(q)} P(z),\qquad
g(q) \;=\; \frac{1}{\pi(q)} \sum_{z\in\mathcal{Z}(q)} P(z)\,g(z),
\end{equation}
where $P(z)$ is the (SCM-implied) probability mass on exact $z$.
We apply this aggregation to obtain oracle experimental and observational distributions
conditioned on $q$, and then compute oracle PNS bounds for $q$ by plugging the aggregated
distributions into the Tian--Pearl formulas in the main text.

\paragraph{Sample-based bounds and thresholding.}
To emulate data scarcity, we generate finite observational and experimental samples and compute
empirical frequencies for each exact $z\in\{0,1\}^{d}$. For a mask query $q$, its empirical
distributions are obtained by aggregating counts over $\mathcal{Z}(q)$ (followed by normalization),
mirroring the oracle mixture aggregation above.
We then plug these empirical distributions into the Tian--Pearl formulas to obtain sample-based
bounds $[LB(q),UB(q)]$ for training.
Finally, we filter low-support queries using a threshold on the number of available samples
(or equivalently, on the overlap/support statistics), and train models only on the retained queries.

\newpage
    \subsection{Experiment Settings}
    
    The code is executed on the following hardware configuration:
    \begin{itemize}
        \item \textbf{CPU:} AMD Ryzen 7 5800H with Radeon Graphics @ 3.20 GHz
        \item \textbf{Memory:} 16 GB RAM
        \item \textbf{System Type:} 64-bit operating system
        \item \textbf{Architecture:} x64-based processor
    \end{itemize}
    The number of observational data and experimental data is from 50k to 500k.

    \subsubsection{Machine Learning Configuration}
    
    \paragraph{MLP backbone (Mish).}
    All learning experiments use the same MLP backbone with Mish activations, and train \emph{two} independent regressors: one for the lower bound and one for the upper bound.
    
    \begin{itemize}
        \item \textbf{Input representation:} each query $q \in \{0,1,X\}^{d}$ is encoded by a $3d$-dimensional one-hot vector. For each coordinate $j\in[d]$, we use three channels indicating $q_j\in\{0,1,X\}$, and concatenate them into $\phi(q)\in\{0,1\}^{3d}$.
        \item \textbf{Architecture:} fully-connected MLP
        \[
            3d \rightarrow 64 \rightarrow 32 \rightarrow 16 \rightarrow 1,
        \]
        with Mish after each hidden layer and a final Sigmoid to ensure predictions lie in $[0,1]$.
        \item \textbf{Outputs:} two separate models are trained, denoted $\widehat{LB}(q)$ and $\widehat{UB}(q)$.
        \item \textbf{Loss:} mean squared error (MSE) between predictions and test. (trained separately for LB and UB).
        \item \textbf{Optimiser:} Adam with learning rate $10^{-3}$.
        \item \textbf{Train/validation split:} $90\%/10\%$ random split; we select the best checkpoint by validation MSE.
        \item \textbf{Training epochs:} up to 2000 epochs; validation is evaluated periodically and the best checkpoint is saved.
        \item \textbf{Batch size:} 4096.
    \end{itemize}
    
    \paragraph{Training protocols}
    We implement three evaluation settings, using the same MLP backbone and training data construction as in the main text:
    \begin{itemize}
        \item \textbf{MLP for Exact Subpopulation (Exact-MLP).}
        Train on thresholded \emph{exact} queries only, i.e., $q\in\{0,1\}^{d}$ (no masked entries).
        Evaluate on the oracle bounds of the exact-query subset.
    
        \item \textbf{Mask-MLP for Exact Subpopulation.}
        Train a single Mask-MLP on the thresholded \emph{mask-augmented} supervision space, i.e., $q\in\{0,1,X\}^{d}$.
        Evaluate on the oracle bounds of the exact-query subset $q\in\{0,1\}^{d}$, enabling direct comparison with MLP for Exact Subpopulation.
    
        \item \textbf{Mask-MLP for Mask-augmented Subpopulation.}
        Using the same Mask-MLP trained above, evaluate on the oracle bounds over the full mask-augmented query space $q\in\{0,1,X\}^{d}$.
    \end{itemize}

\newpage
\subsection{Experimental Result}
\label{app:more_plots}

This appendix reports additional diagnostic plots for all budgets.
For each budget, we include (i) threshold sweeps of MAE (each PDF contains two panels: exact queries on the left and mask-augmented queries on the right),
and (ii) qualitative true-vs-predicted plots on oracle bounds.

\subsubsection{Budget = 50k}
\label{app:plots_50k}
\begin{figure}[H]
    \centering
    \begin{subfigure}[b]{0.49\linewidth}
        \centering
        \includegraphics[width=\linewidth]{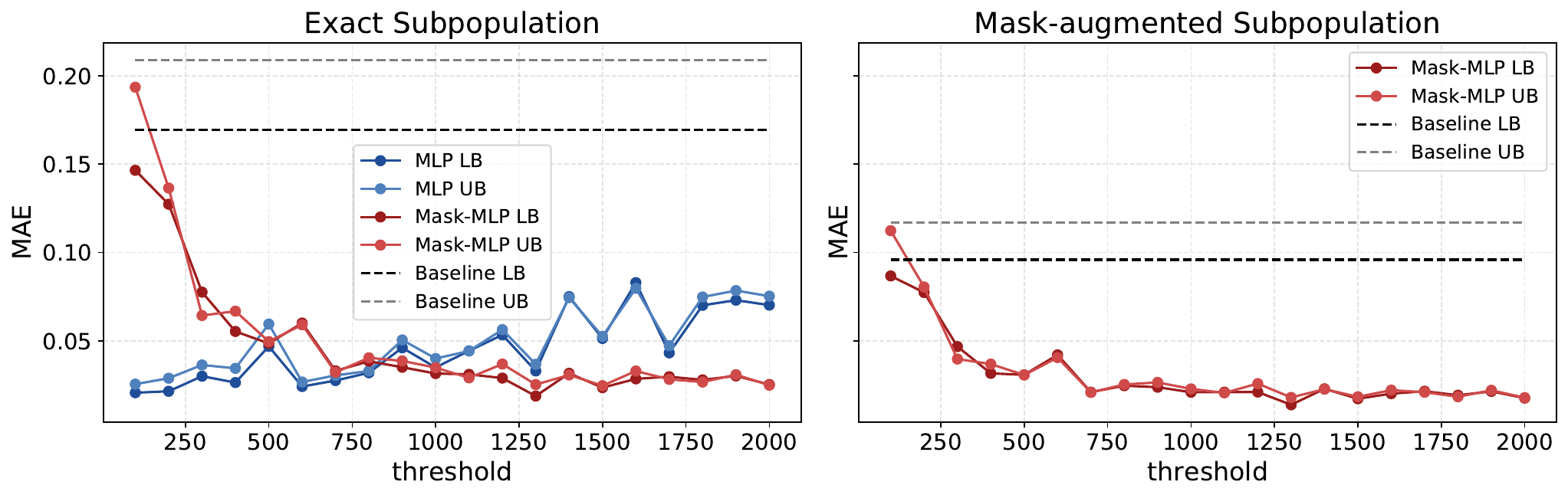}
        \caption{Direct}
        \label{fig:app_mae_direct_50k}
    \end{subfigure}
    \hfill
    \begin{subfigure}[b]{0.49\linewidth}
        \centering
        \includegraphics[width=\linewidth]{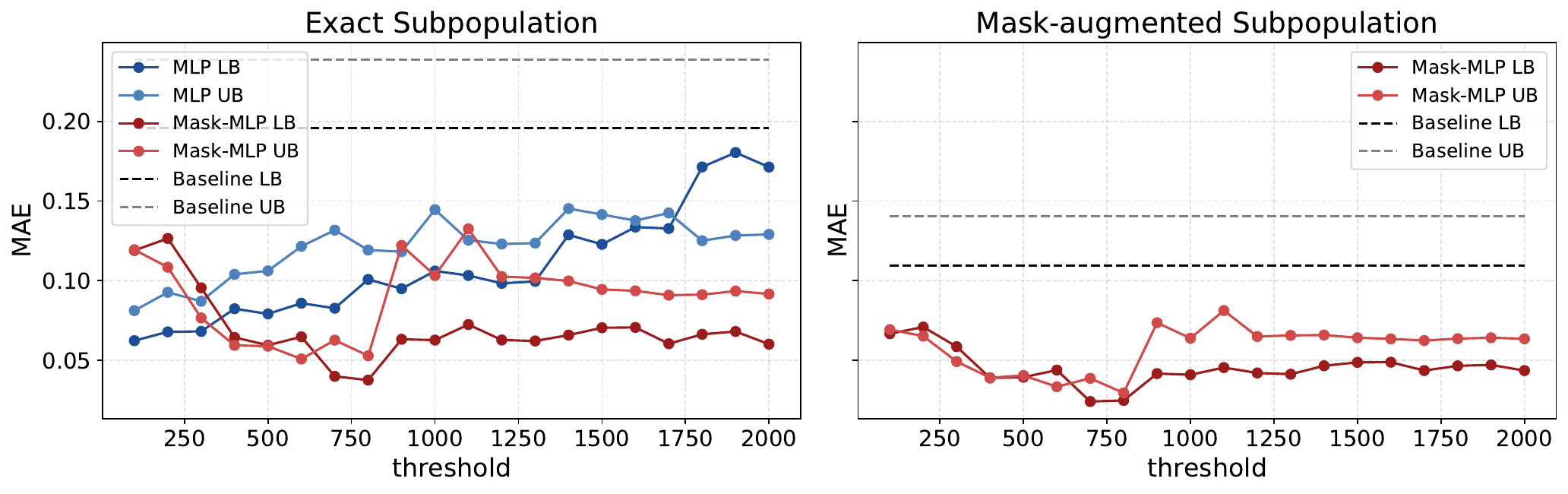}
        \caption{Covariate}
        \label{fig:app_mae_cov_50k}
    \end{subfigure}

    \vspace{2mm}

    \begin{subfigure}[b]{0.49\linewidth}
        \centering
        \includegraphics[width=\linewidth]{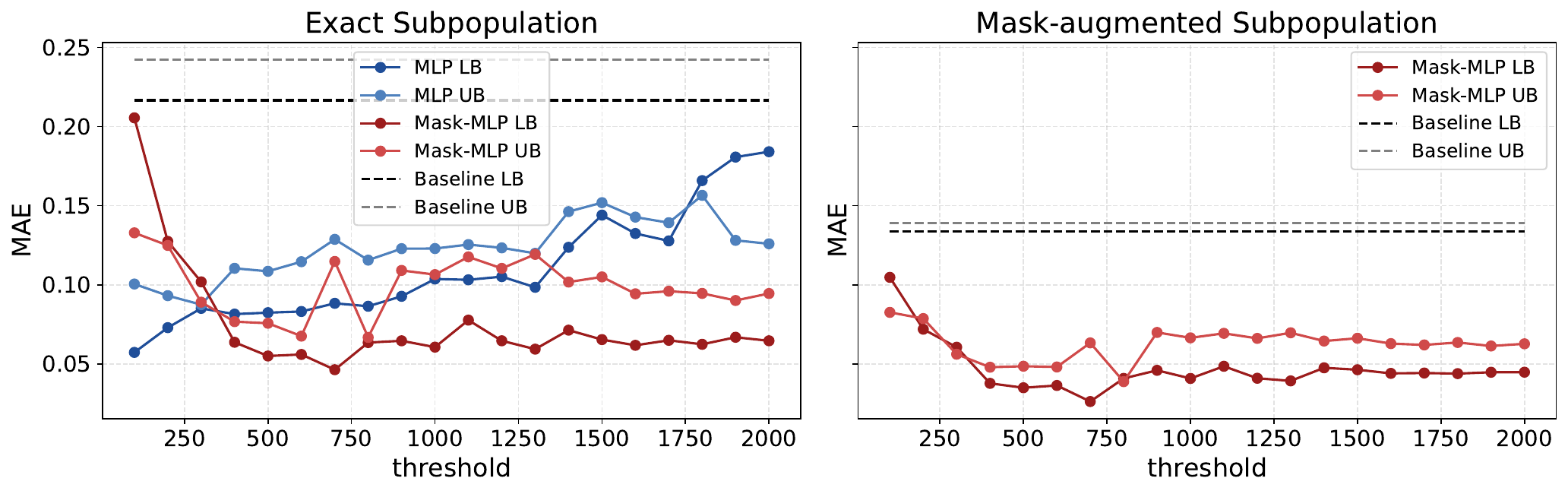}
        \caption{Confounder}
        \label{fig:app_mae_conf_50k}
    \end{subfigure}
    \hfill
    \begin{subfigure}[b]{0.49\linewidth}
        \centering
        \includegraphics[width=\linewidth]{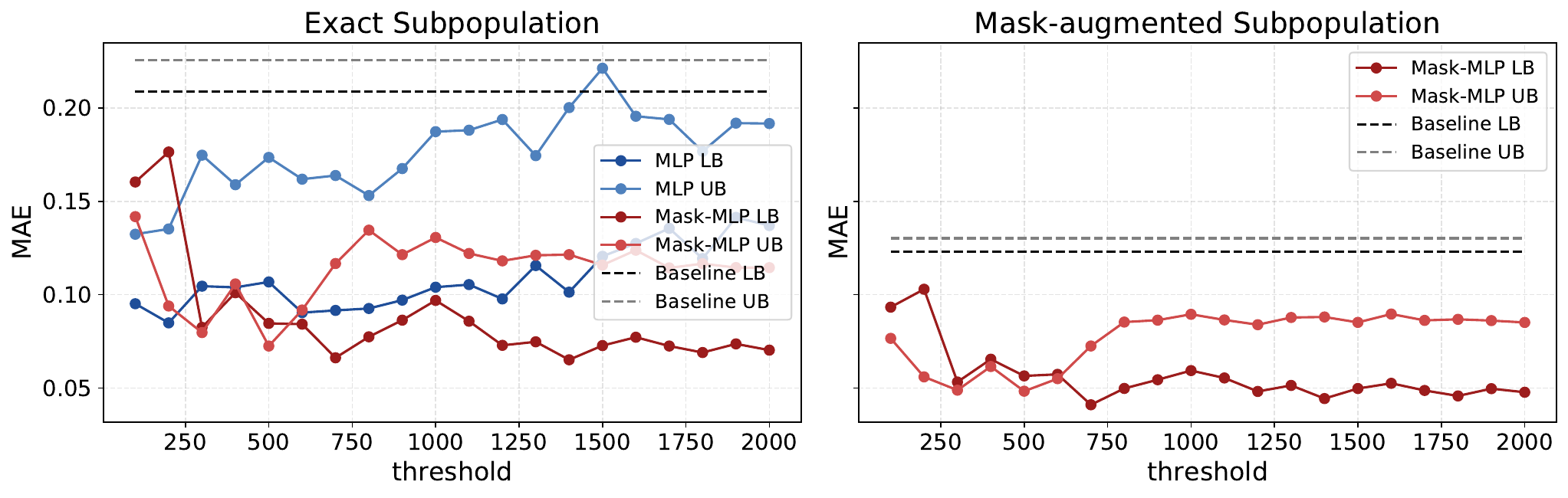}
        \caption{Mediator}
        \label{fig:app_mae_med_50k}
    \end{subfigure}
    \caption{Threshold sweeps (50k budget). Each PDF contains two panels: evaluation on exact queries (left) and on mask-augmented queries (right).}
    \label{fig:app_mae_sweep_50k}
\end{figure}

\begin{figure}[H]
    \centering
    \begin{subfigure}[b]{0.49\linewidth}
        \centering
        \includegraphics[width=\linewidth]{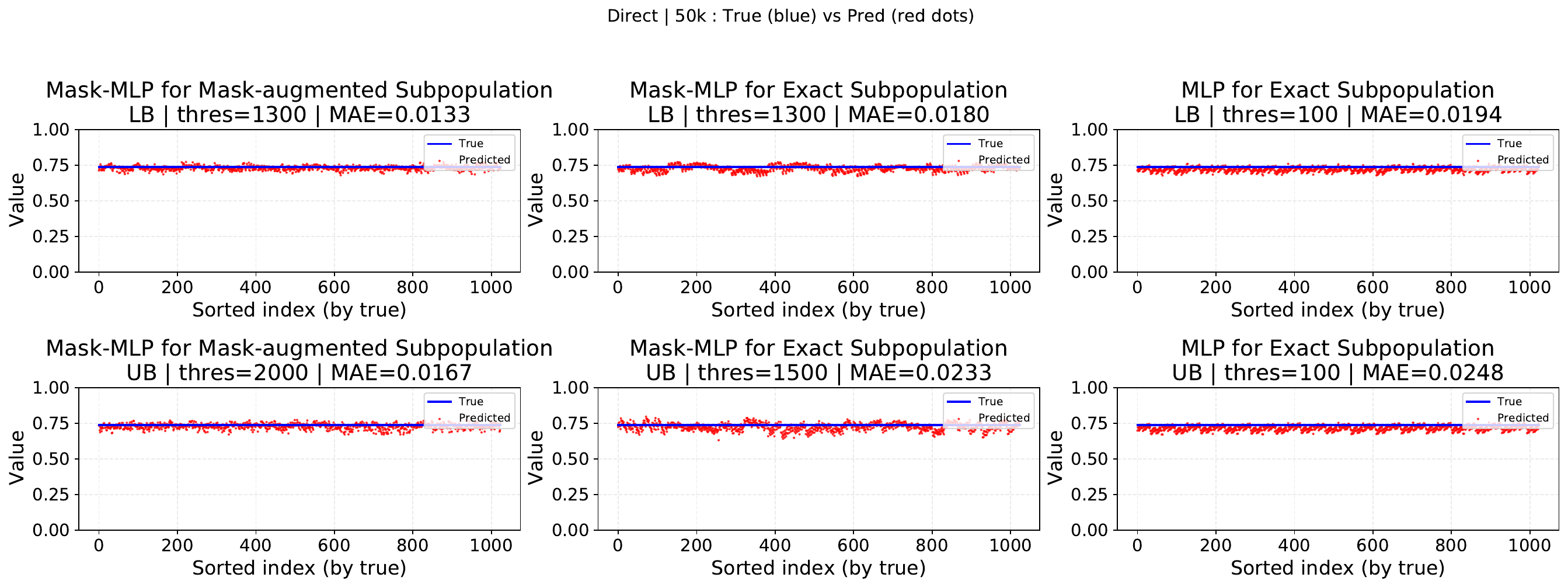}
        \caption{Direct}
        \label{fig:app_tp_direct_50k}
    \end{subfigure}
    \hfill
    \begin{subfigure}[b]{0.49\linewidth}
        \centering
        \includegraphics[width=\linewidth]{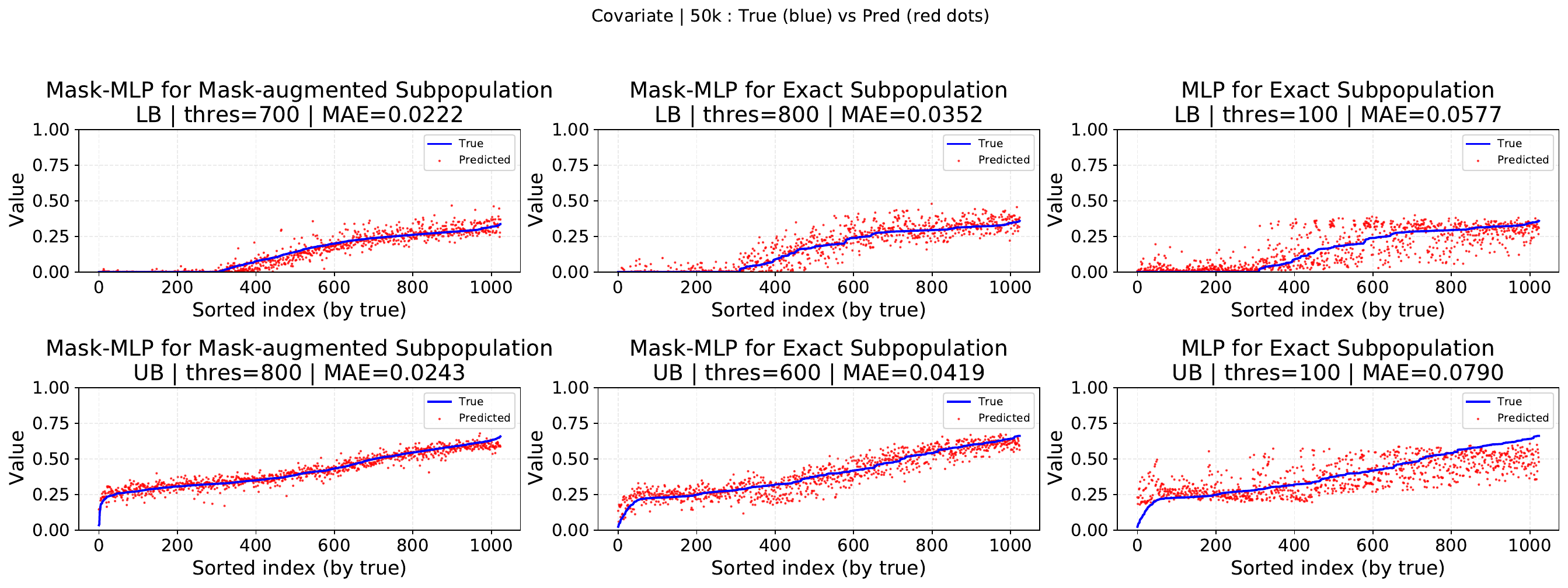}
        \caption{Covariate}
        \label{fig:app_tp_cov_50k}
    \end{subfigure}

    \vspace{2mm}

    \begin{subfigure}[b]{0.49\linewidth}
        \centering
        \includegraphics[width=\linewidth]{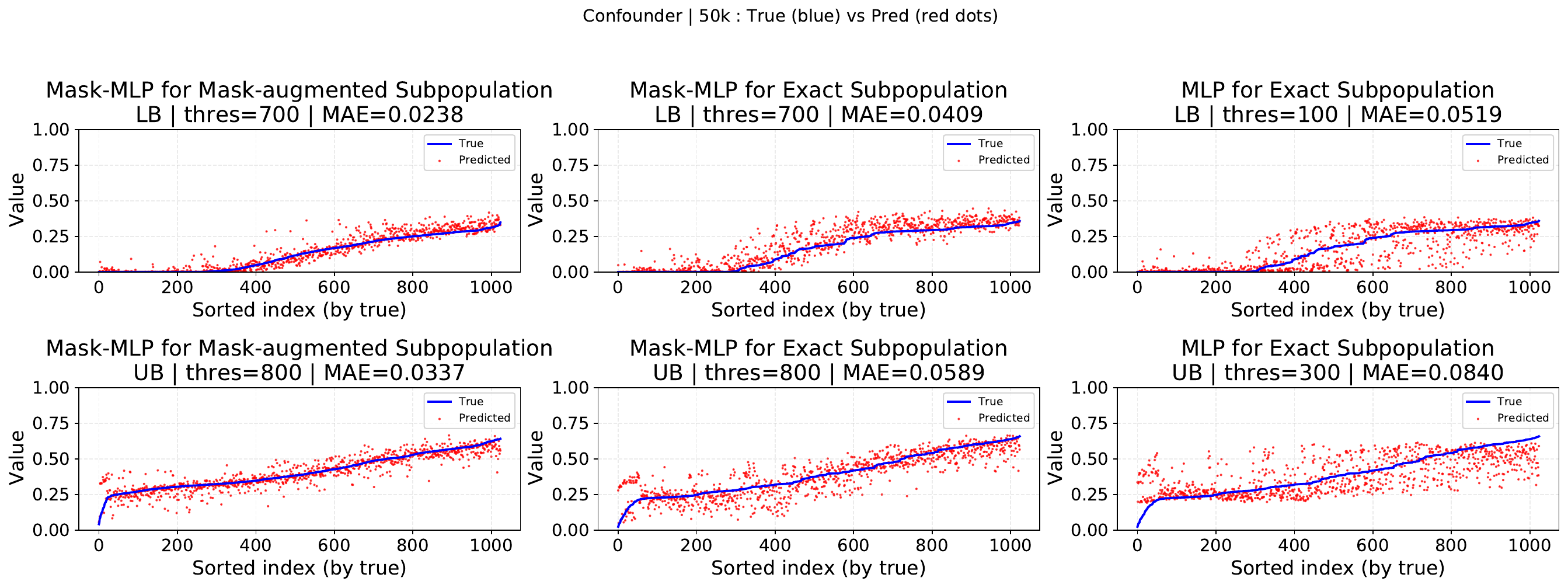}
        \caption{Confounder}
        \label{fig:app_tp_conf_50k}
    \end{subfigure}
    \hfill
    \begin{subfigure}[b]{0.49\linewidth}
        \centering
        \includegraphics[width=\linewidth]{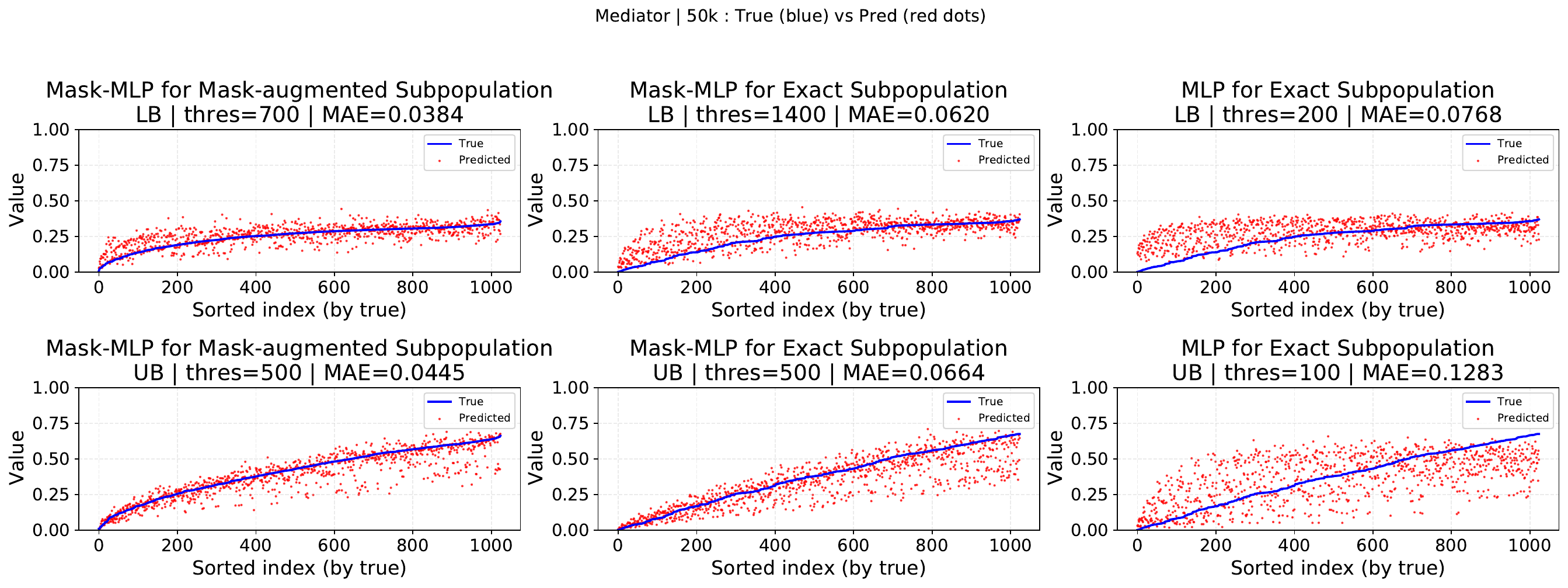}
        \caption{Mediator}
        \label{fig:app_tp_med_50k}
    \end{subfigure}
    \caption{True vs.\ predicted oracle bounds (50k budget). Each PDF visualizes prediction fidelity; see legends within each plot for exact vs.\ mask-augmented evaluations.}
    \label{fig:app_truepred_50k}
\end{figure}

\subsubsection{Budget = 100k}
\label{app:plots_100k}

\begin{figure}[H]
    \centering
    \begin{subfigure}[b]{0.49\linewidth}
        \centering
        \includegraphics[width=\linewidth]{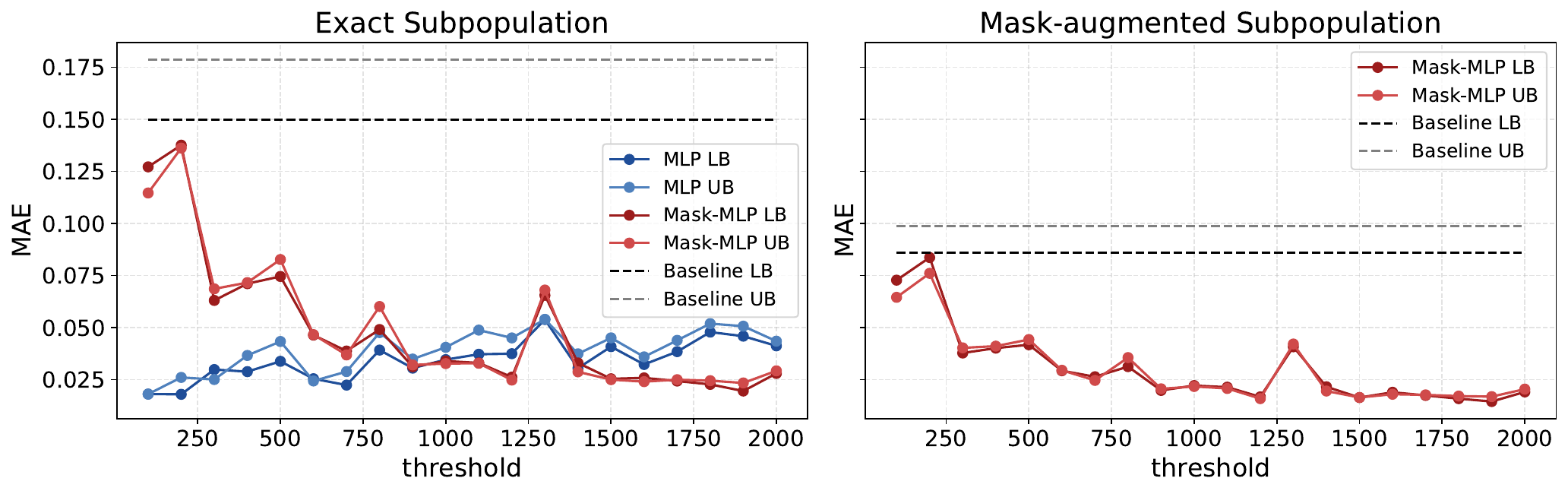}
        \caption{Direct}
        \label{fig:app_mae_direct_100k}
    \end{subfigure}
    \hfill
    \begin{subfigure}[b]{0.49\linewidth}
        \centering
        \includegraphics[width=\linewidth]{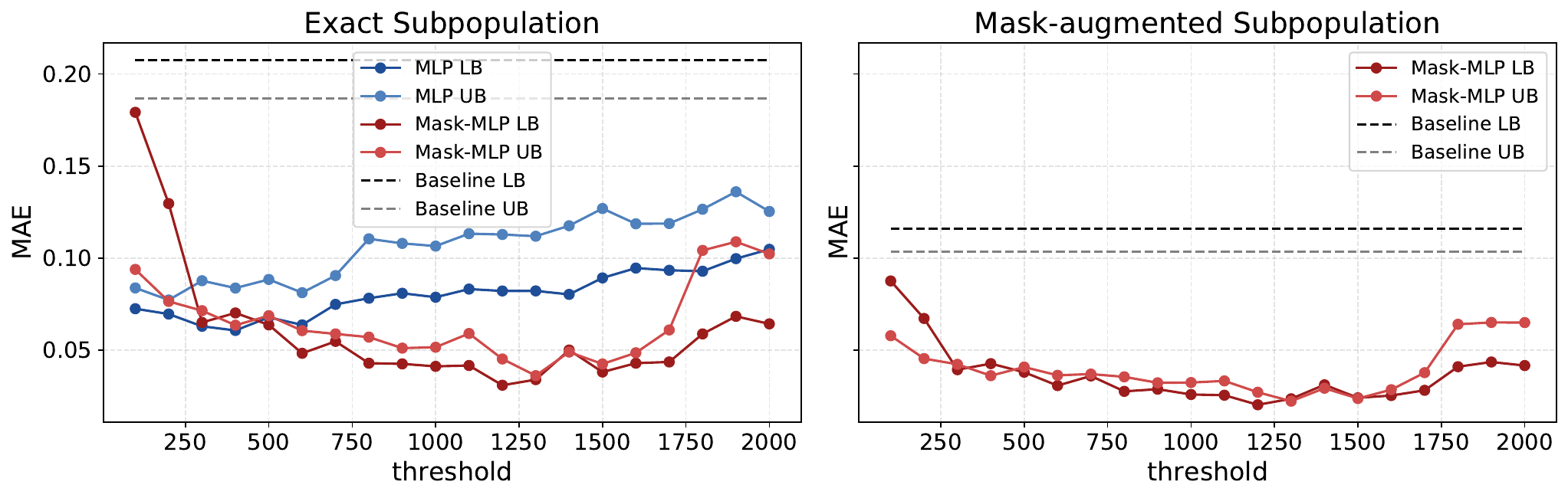}
        \caption{Covariate}
        \label{fig:app_mae_cov_100k}
    \end{subfigure}

    \vspace{2mm}

    \begin{subfigure}[b]{0.49\linewidth}
        \centering
        \includegraphics[width=\linewidth]{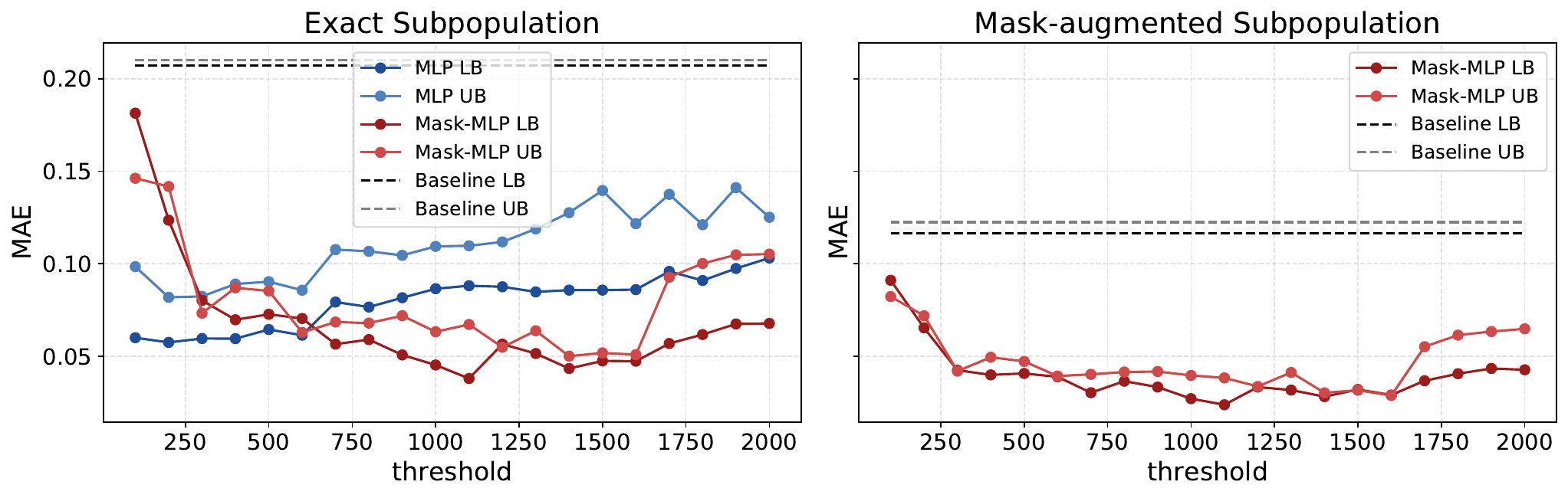}
        \caption{Confounder}
        \label{fig:app_mae_conf_100k}
    \end{subfigure}
    \hfill
    \begin{subfigure}[b]{0.49\linewidth}
        \centering
        \includegraphics[width=\linewidth]{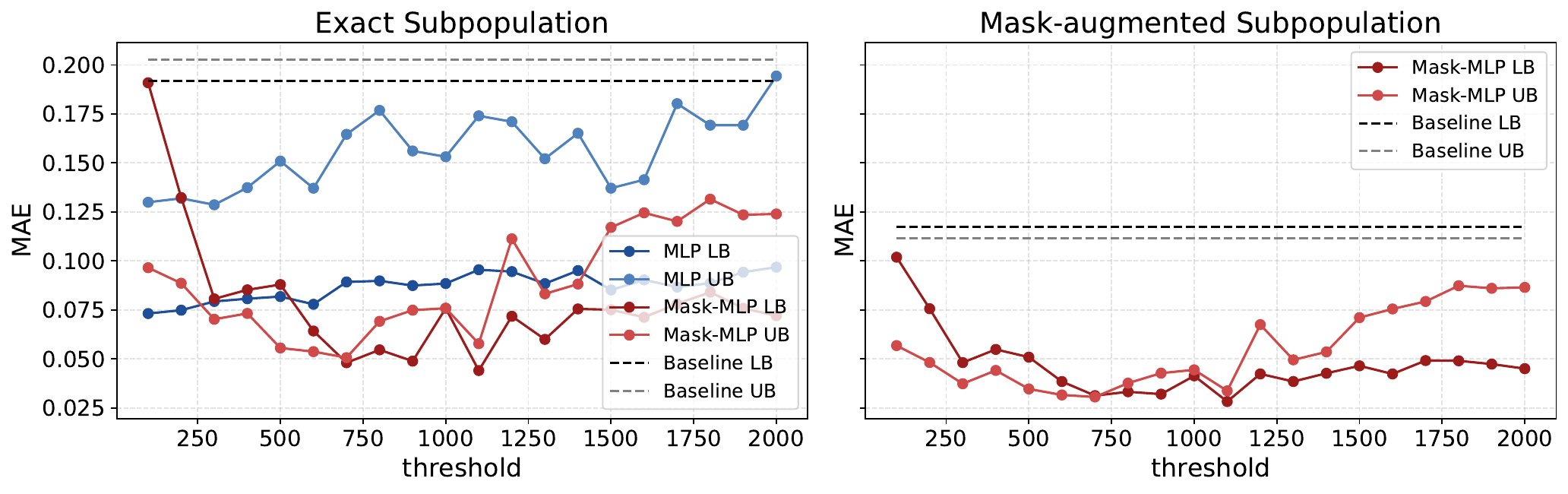}
        \caption{Mediator}
        \label{fig:app_mae_med_100k}
    \end{subfigure}
    \caption{Threshold sweeps (100k budget). Each PDF contains two panels: evaluation on exact queries (left) and on mask-augmented queries (right).}
    \label{fig:app_mae_sweep_100k}
\end{figure}

\begin{figure}[H]
    \centering
    \begin{subfigure}[b]{0.49\linewidth}
        \centering
        \includegraphics[width=\linewidth]{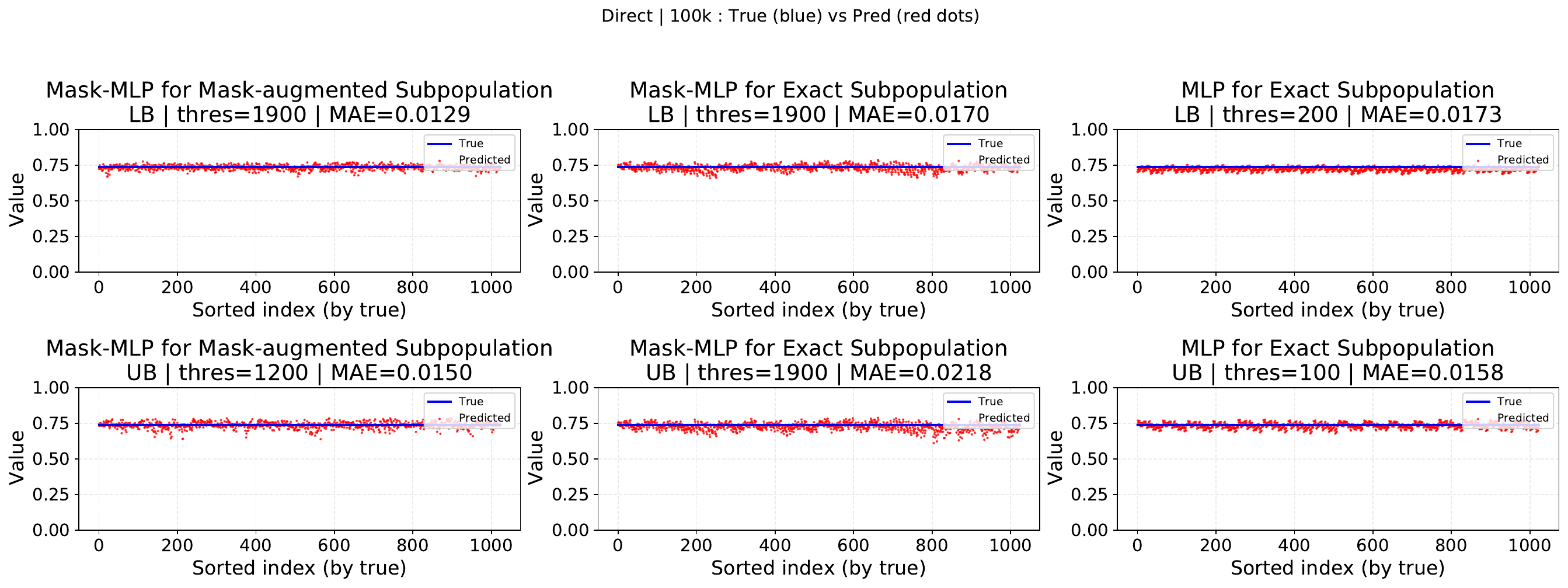}
        \caption{Direct}
        \label{fig:app_tp_direct_100k}
    \end{subfigure}
    \hfill
    \begin{subfigure}[b]{0.49\linewidth}
        \centering
        \includegraphics[width=\linewidth]{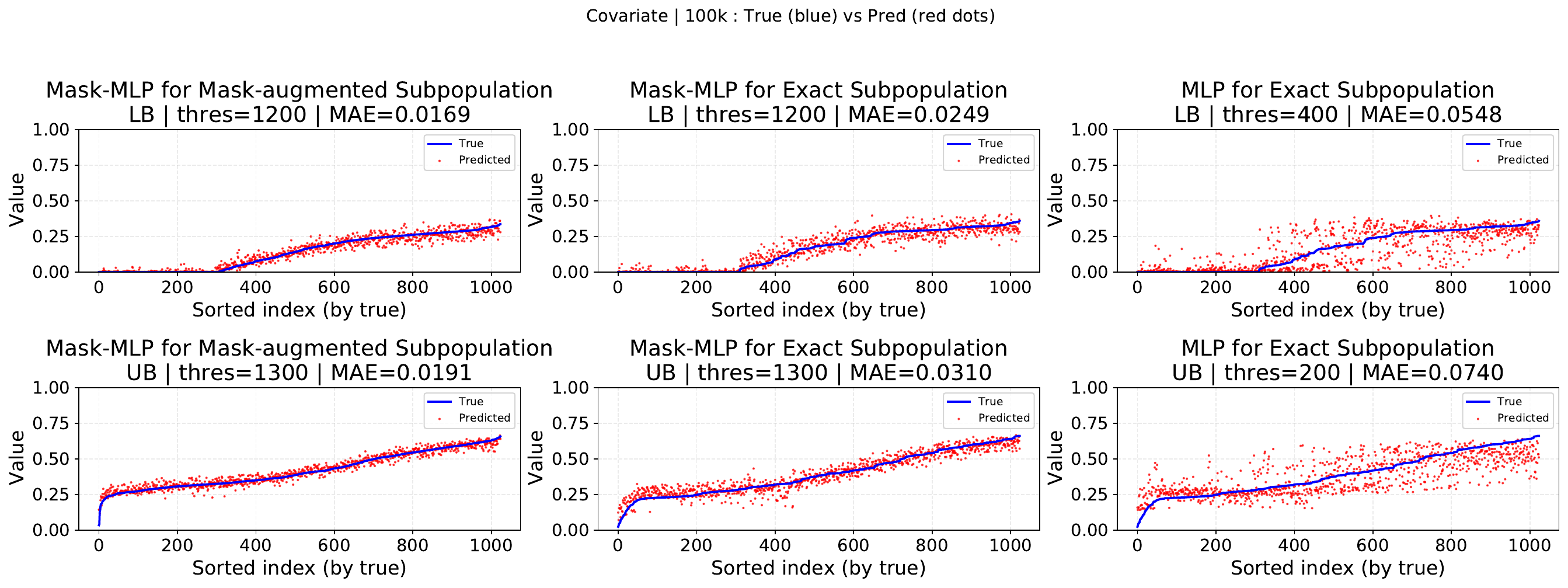}
        \caption{Covariate}
        \label{fig:app_tp_cov_100k}
    \end{subfigure}

    \vspace{2mm}

    \begin{subfigure}[b]{0.49\linewidth}
        \centering
        \includegraphics[width=\linewidth]{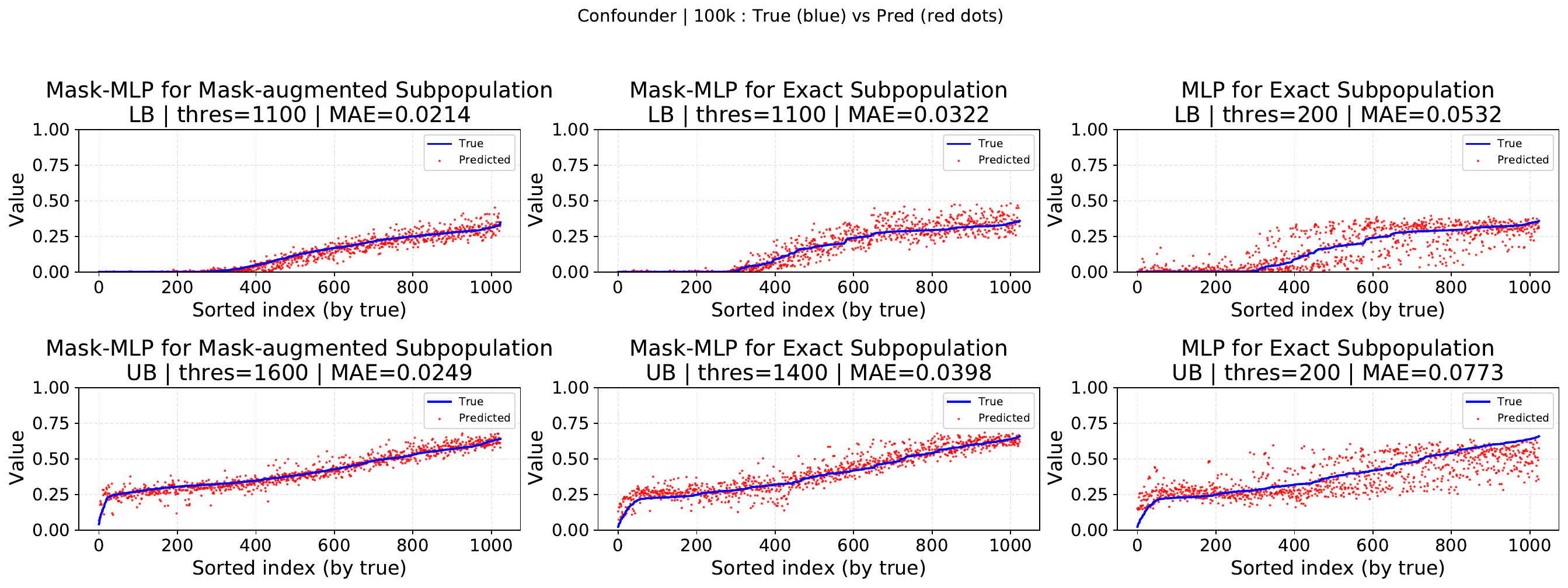}
        \caption{Confounder}
        \label{fig:app_tp_conf_100k}
    \end{subfigure}
    \hfill
    \begin{subfigure}[b]{0.49\linewidth}
        \centering
        \includegraphics[width=\linewidth]{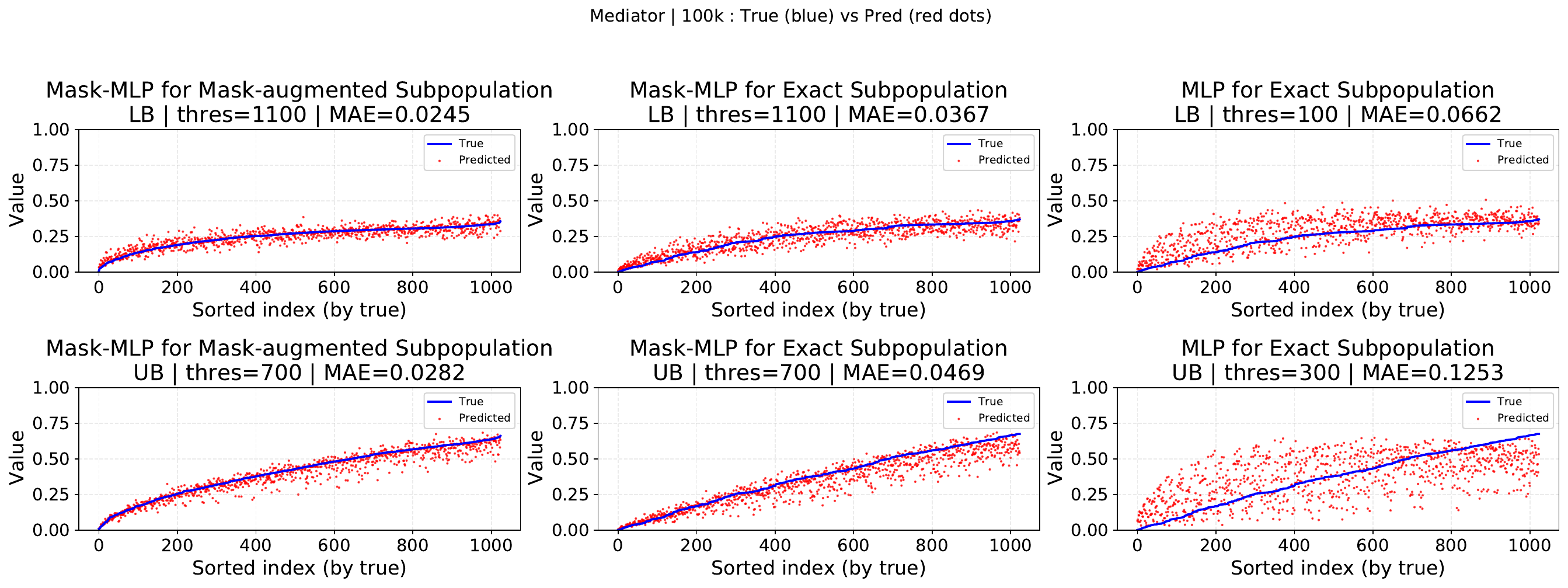}
        \caption{Mediator}
        \label{fig:app_tp_med_100k}
    \end{subfigure}
    \caption{True vs.\ predicted oracle bounds (100k budget).}
    \label{fig:app_truepred_100k}
\end{figure}

\subsubsection{Budget = 200k}
\label{app:plots_200k}

\begin{figure}[H]
    \centering
    \begin{subfigure}[b]{0.49\linewidth}
        \centering
        \includegraphics[width=\linewidth]{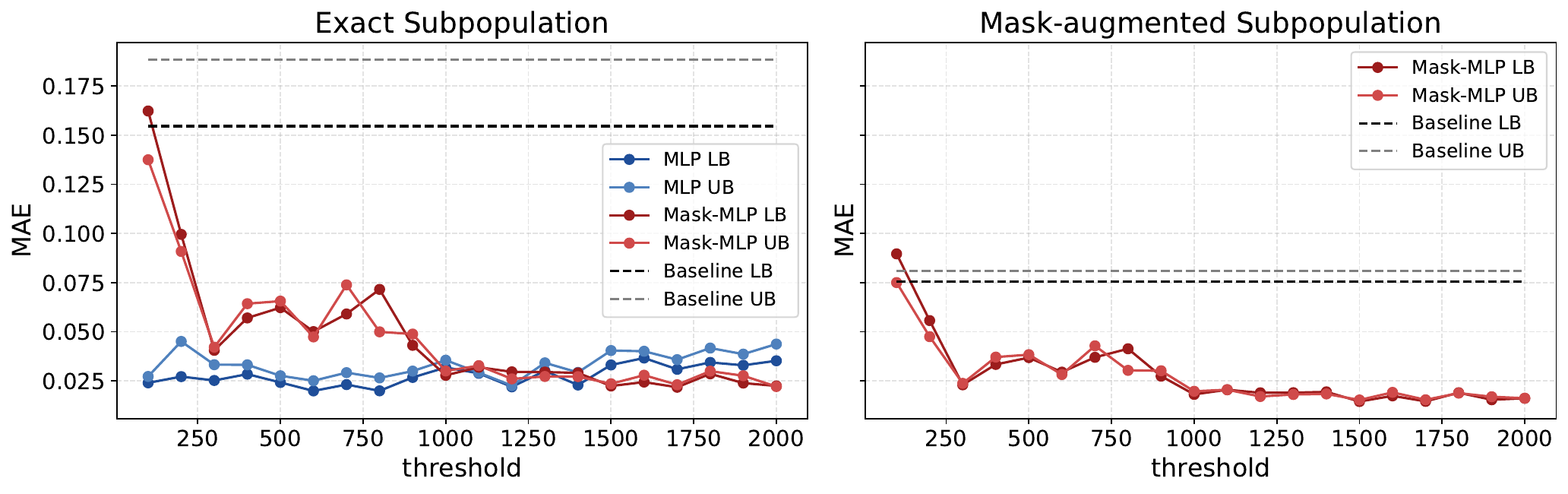}
        \caption{Direct}
        \label{fig:app_mae_direct_200k}
    \end{subfigure}
    \hfill
    \begin{subfigure}[b]{0.49\linewidth}
        \centering
        \includegraphics[width=\linewidth]{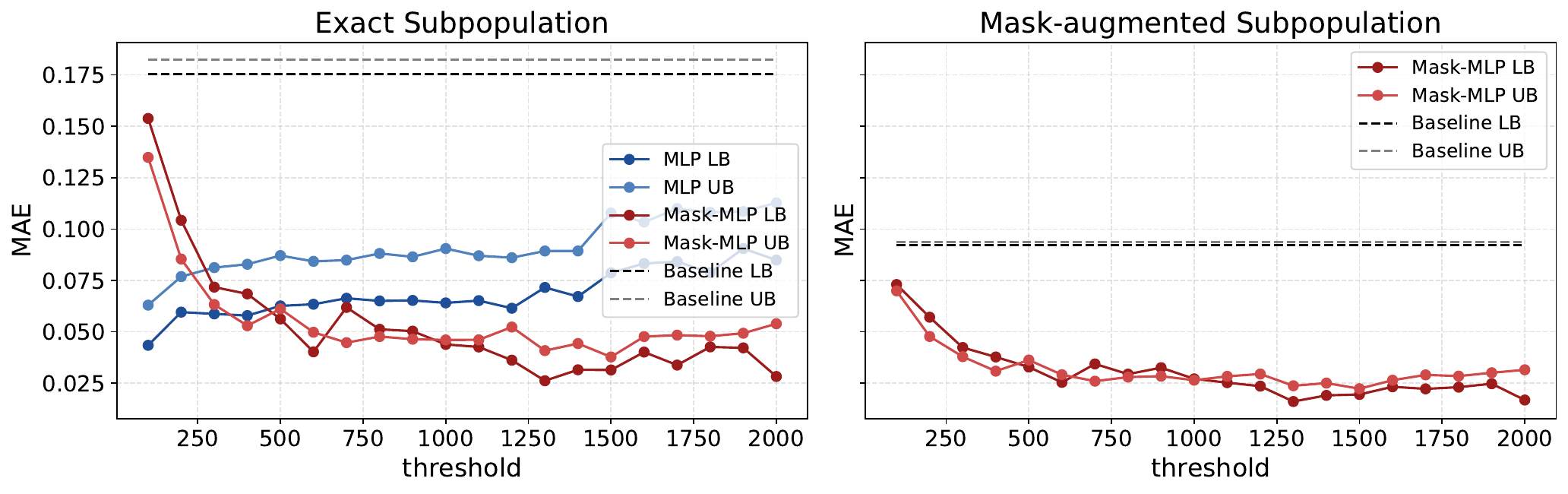}
        \caption{Covariate}
        \label{fig:app_mae_cov_200k}
    \end{subfigure}

    \vspace{2mm}

    \begin{subfigure}[b]{0.49\linewidth}
        \centering
        \includegraphics[width=\linewidth]{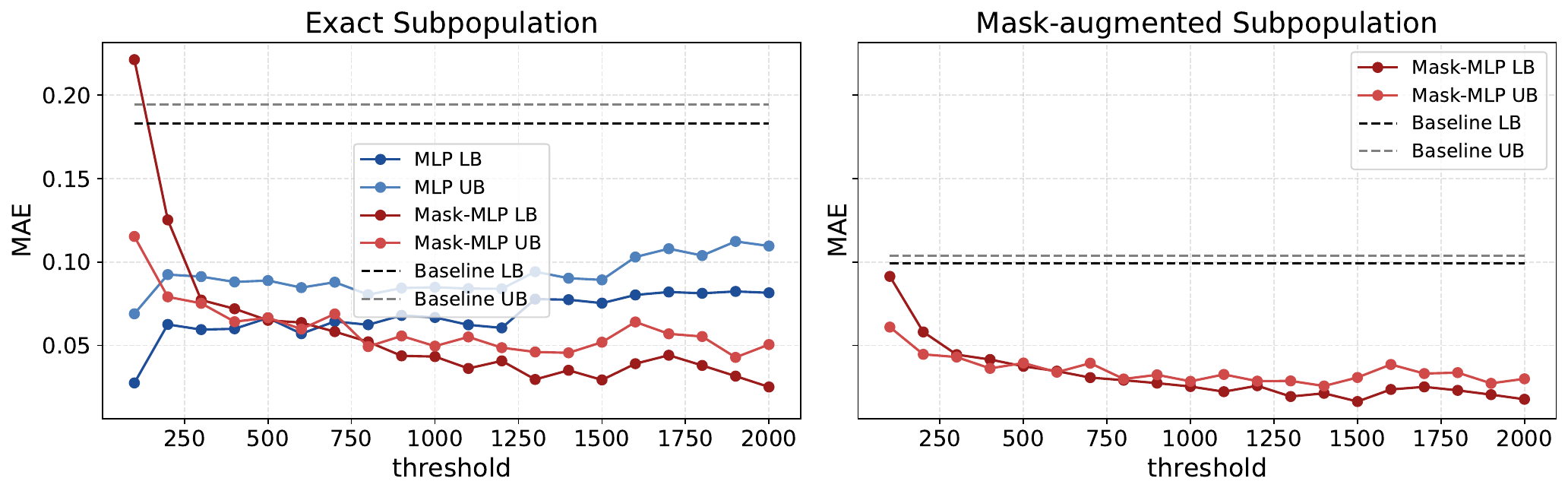}
        \caption{Confounder}
        \label{fig:app_mae_conf_200k}
    \end{subfigure}
    \hfill
    \begin{subfigure}[b]{0.49\linewidth}
        \centering
        \includegraphics[width=\linewidth]{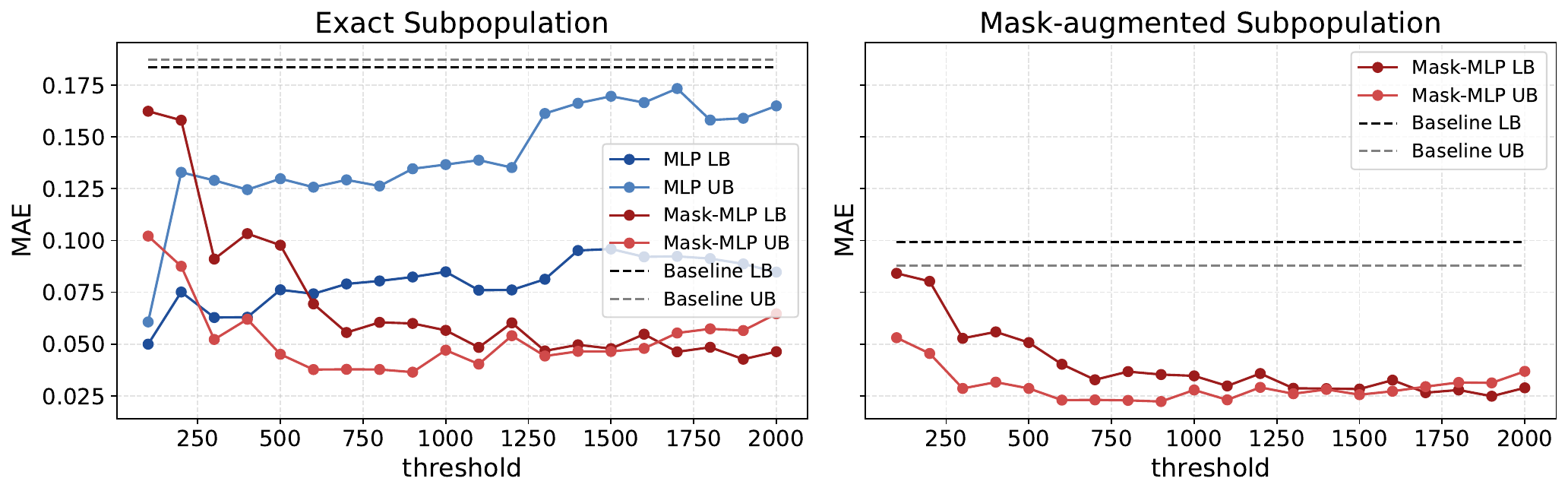}
        \caption{Mediator}
        \label{fig:app_mae_med_200k}
    \end{subfigure}
    \caption{Threshold sweeps (200k budget). Each PDF contains two panels: evaluation on exact queries (left) and on mask-augmented queries (right).}
    \label{fig:app_mae_sweep_200k}
\end{figure}

\begin{figure}[H]
    \centering
    \begin{subfigure}[b]{0.49\linewidth}
        \centering
        \includegraphics[width=\linewidth]{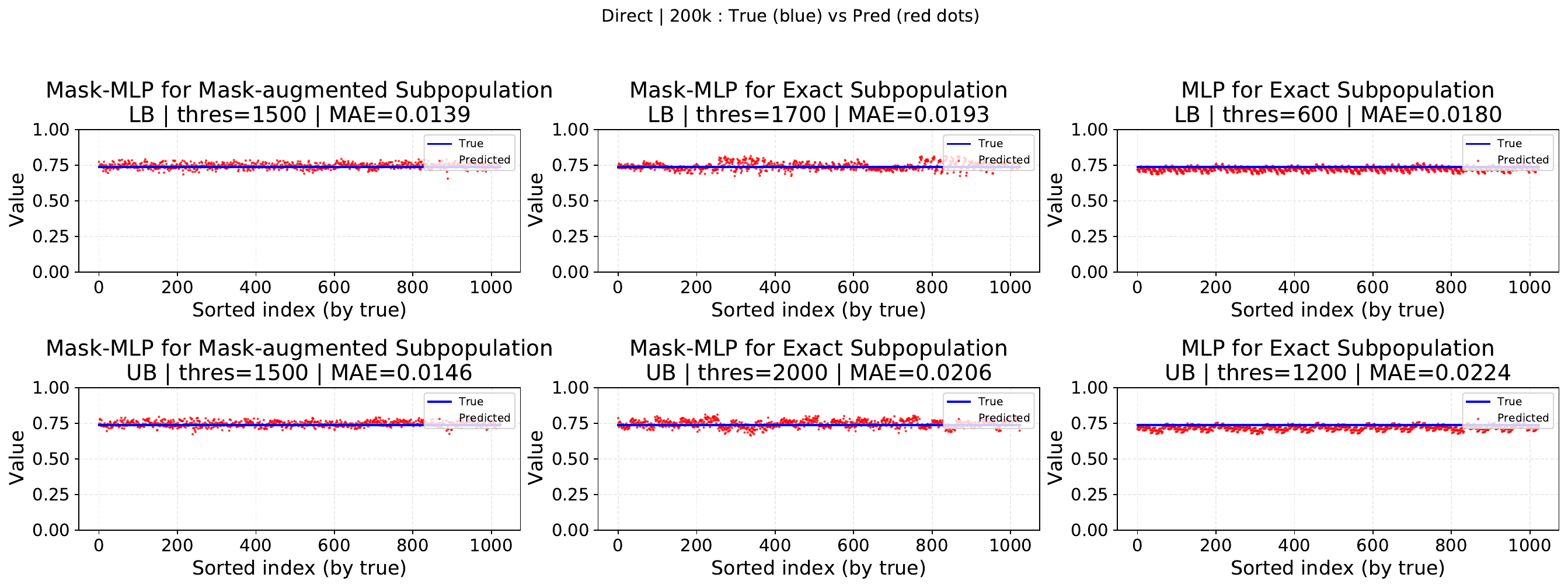}
        \caption{Direct}
        \label{fig:app_tp_direct_200k}
    \end{subfigure}
    \hfill
    \begin{subfigure}[b]{0.49\linewidth}
        \centering
        \includegraphics[width=\linewidth]{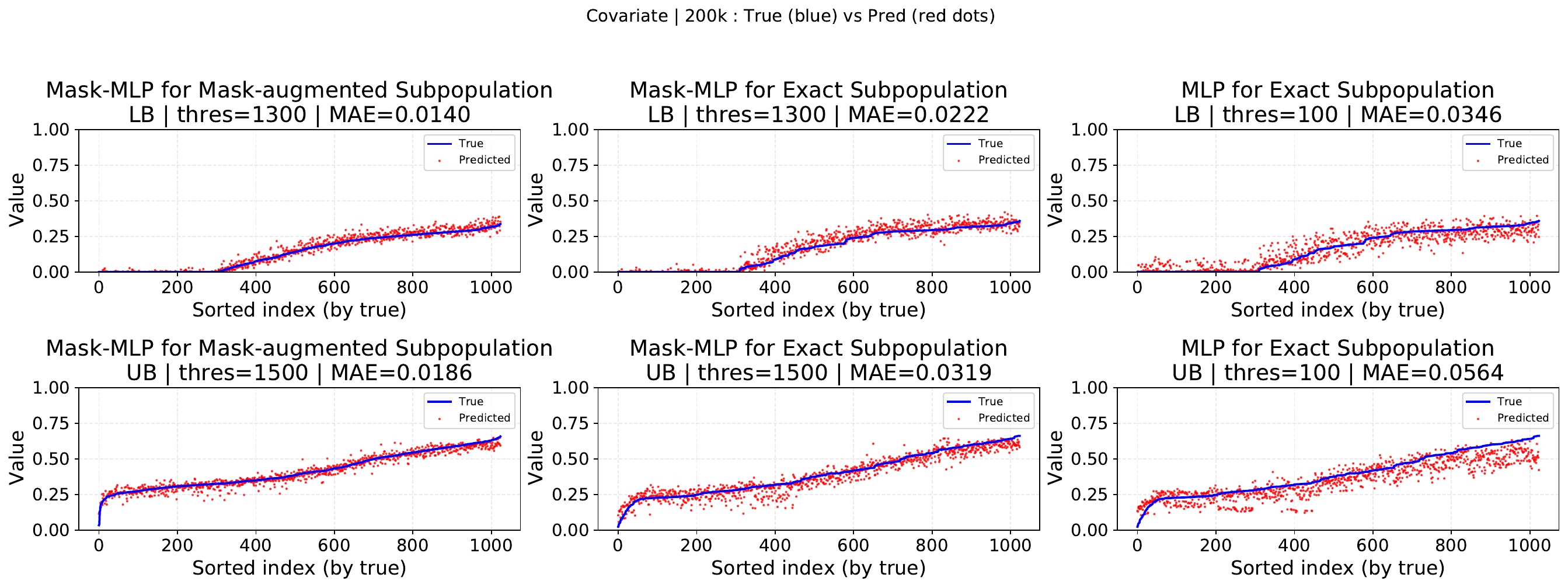}
        \caption{Covariate}
        \label{fig:app_tp_cov_200k}
    \end{subfigure}

    \vspace{2mm}

    \begin{subfigure}[b]{0.49\linewidth}
        \centering
        \includegraphics[width=\linewidth]{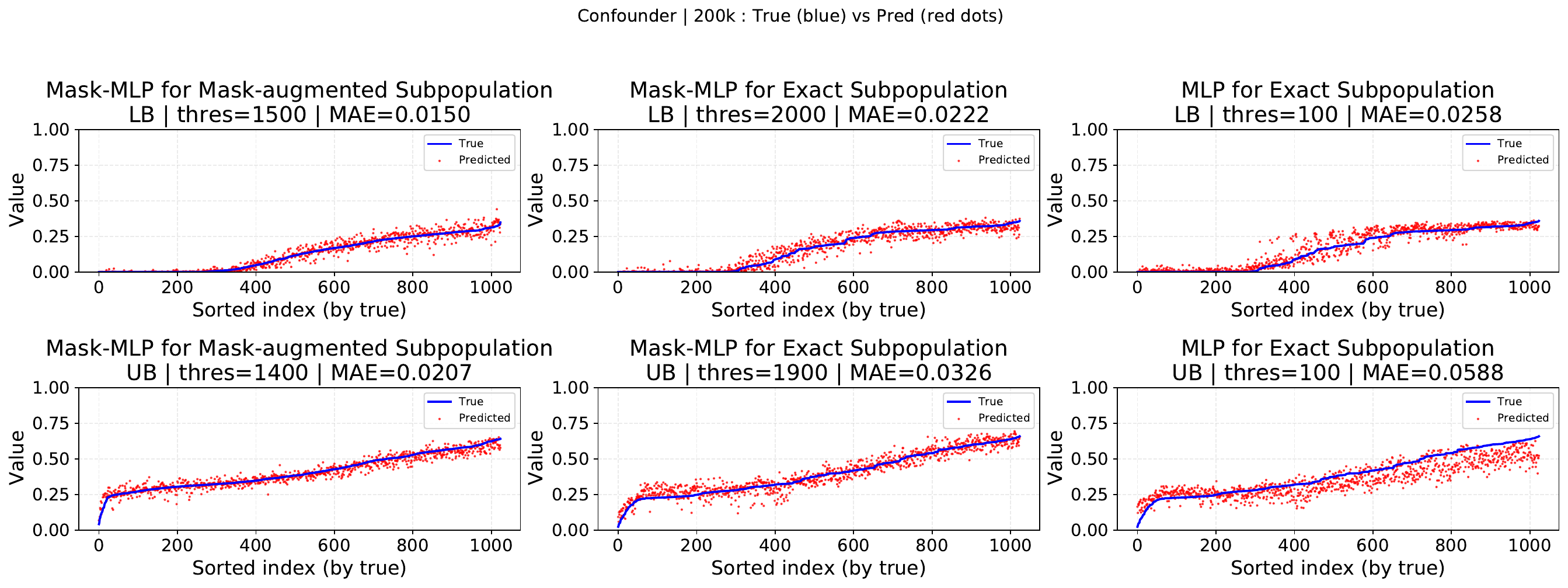}
        \caption{Confounder}
        \label{fig:app_tp_conf_200k}
    \end{subfigure}
    \hfill
    \begin{subfigure}[b]{0.49\linewidth}
        \centering
        \includegraphics[width=\linewidth]{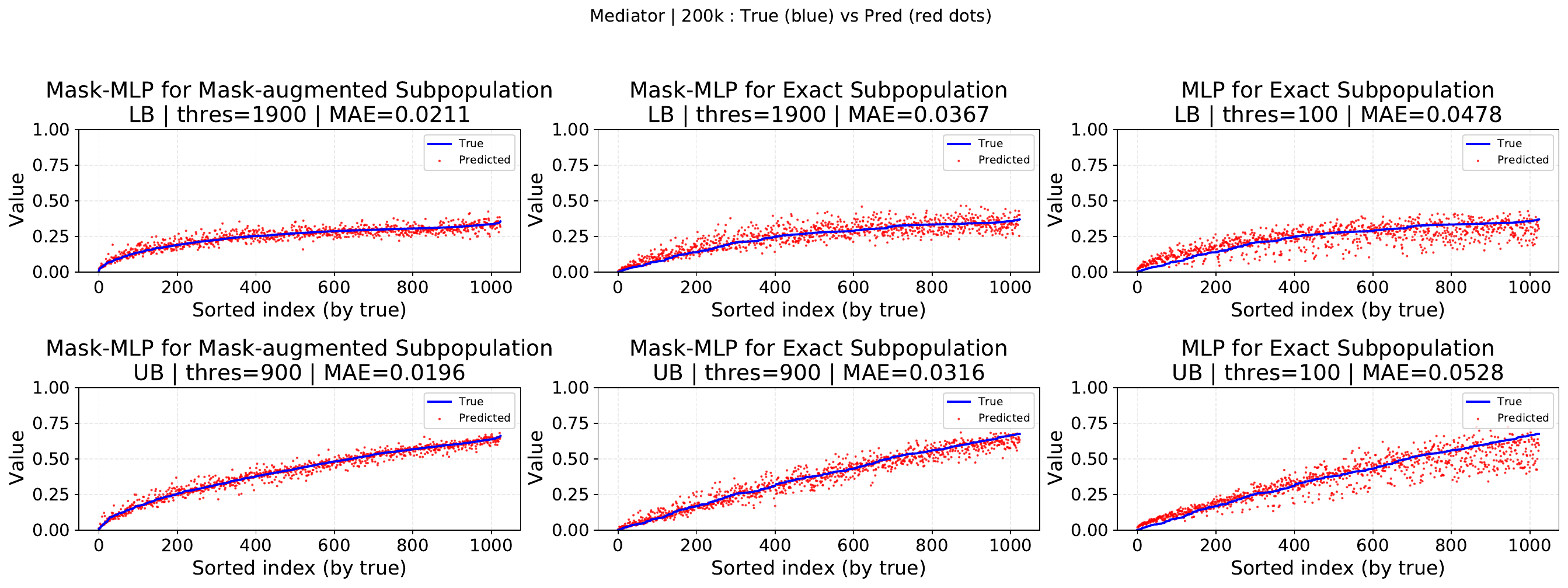}
        \caption{Mediator}
        \label{fig:app_tp_med_200k}
    \end{subfigure}
    \caption{True vs.\ predicted oracle bounds (200k budget).}
    \label{fig:app_truepred_200k}
\end{figure}

\subsubsection{Budget = 500k}
\label{app:plots_500k}

\begin{figure}[H]
    \centering
    \begin{subfigure}[b]{0.49\linewidth}
        \centering
        \includegraphics[width=\linewidth]{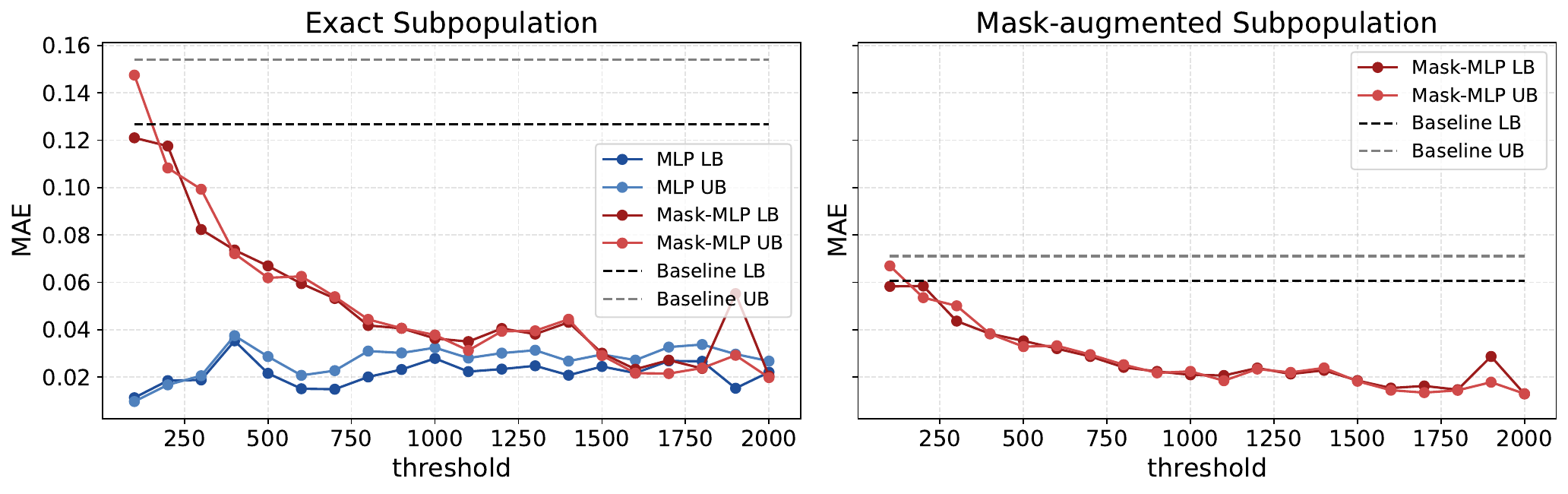}
        \caption{Direct}
        \label{fig:app_mae_direct_500k}
    \end{subfigure}
    \hfill
    \begin{subfigure}[b]{0.49\linewidth}
        \centering
        \includegraphics[width=\linewidth]{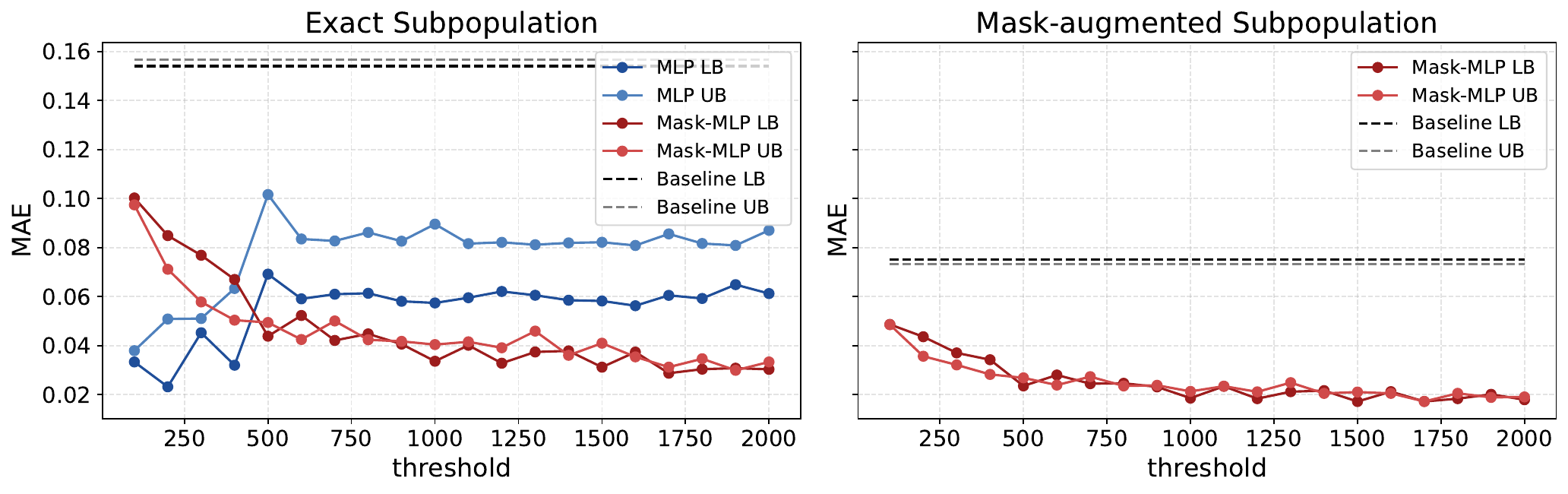}
        \caption{Covariate}
        \label{fig:app_mae_cov_500k}
    \end{subfigure}

    \vspace{2mm}

    \begin{subfigure}[b]{0.49\linewidth}
        \centering
        \includegraphics[width=\linewidth]{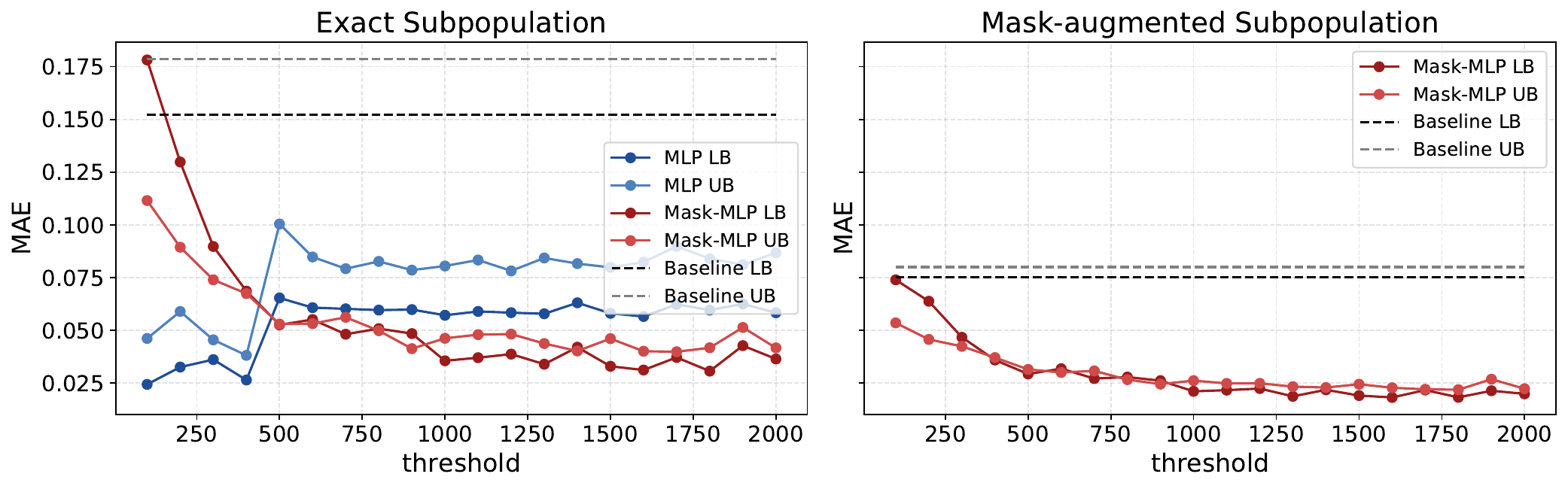}
        \caption{Confounder}
        \label{fig:app_mae_conf_500k}
    \end{subfigure}
    \hfill
    \begin{subfigure}[b]{0.49\linewidth}
        \centering
        \includegraphics[width=\linewidth]{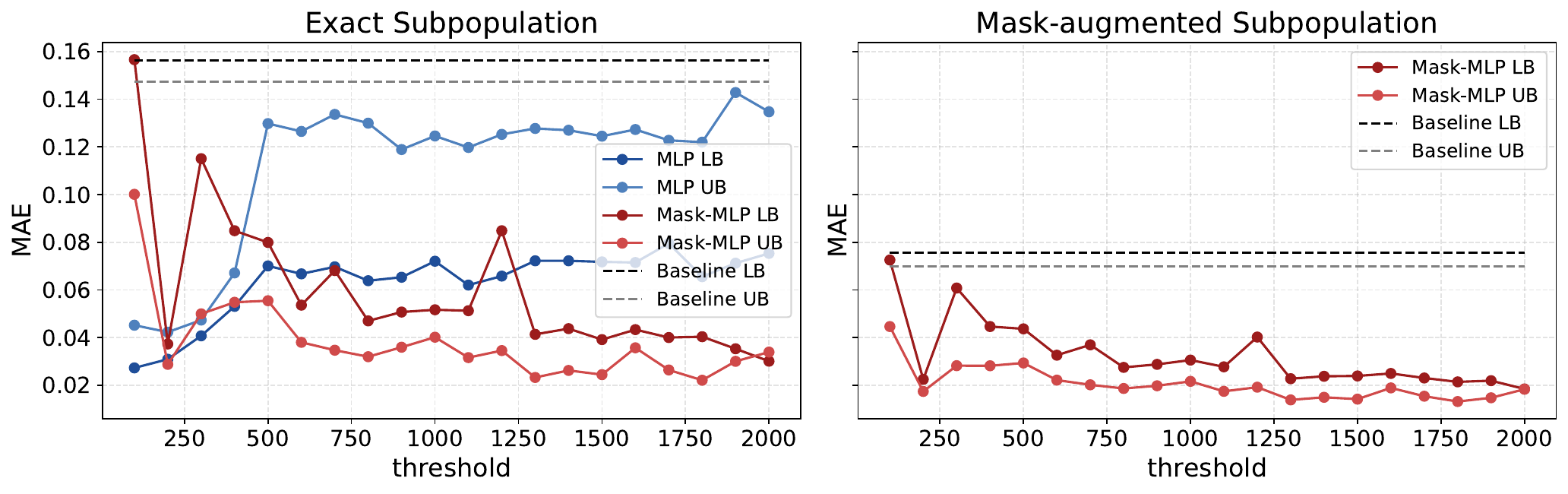}
        \caption{Mediator}
        \label{fig:app_mae_med_500k}
    \end{subfigure}
    \caption{Threshold sweeps (500k budget). Each PDF contains two panels: evaluation on exact queries (left) and on mask-augmented queries (right).}
    \label{fig:app_mae_sweep_500k}
\end{figure}

\begin{figure}[H]
    \centering
    \begin{subfigure}[b]{0.49\linewidth}
        \centering
        \includegraphics[width=\linewidth]{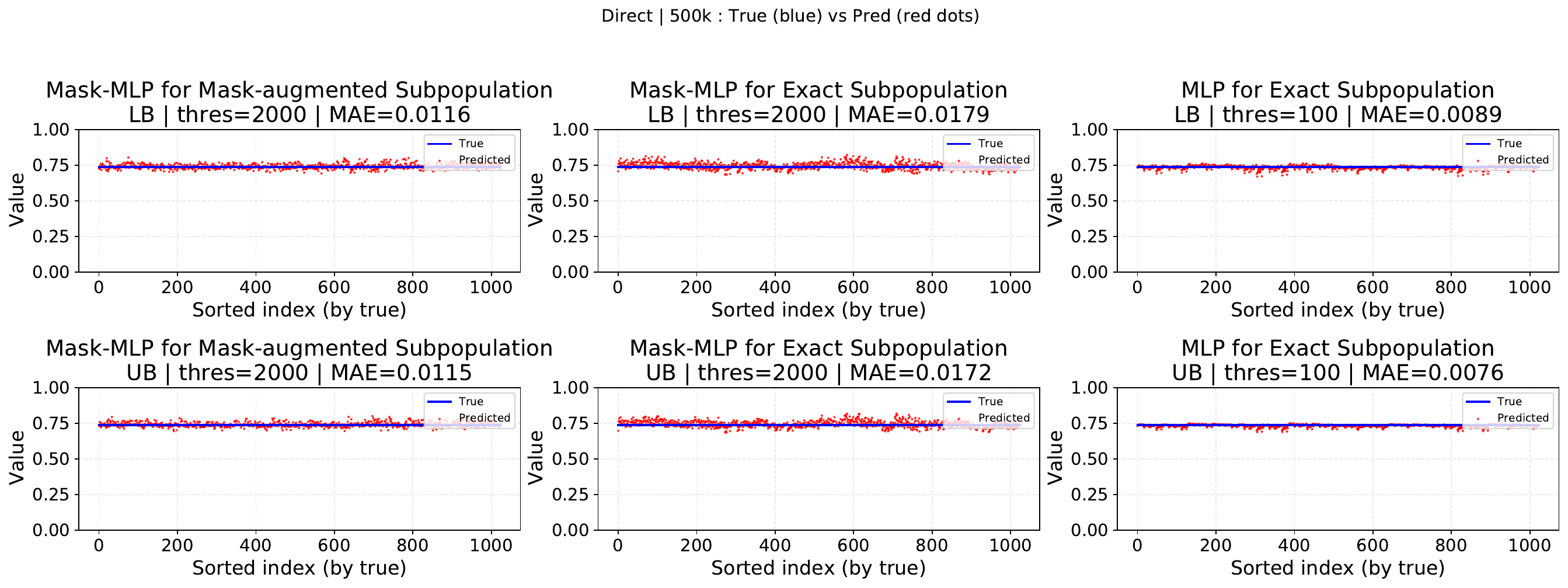}
        \caption{Direct}
        \label{fig:app_tp_direct_500k}
    \end{subfigure}
    \hfill
    \begin{subfigure}[b]{0.49\linewidth}
        \centering
        \includegraphics[width=\linewidth]{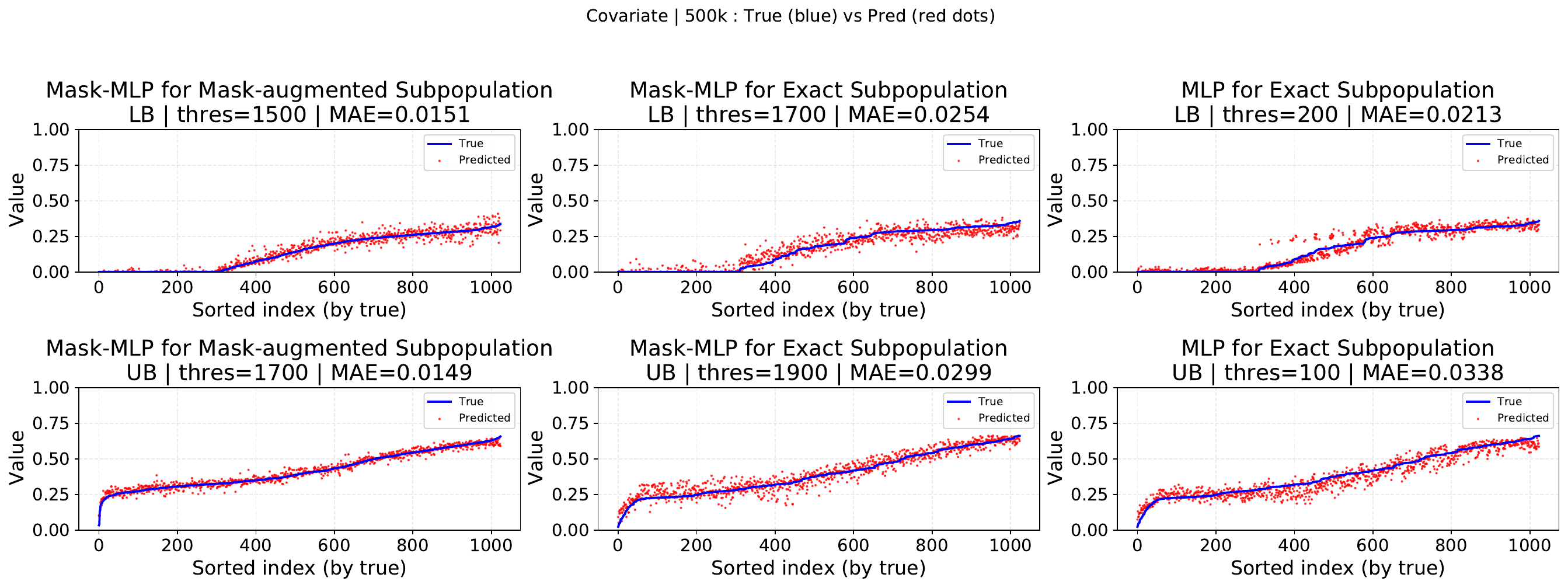}
        \caption{Covariate}
        \label{fig:app_tp_cov_500k}
    \end{subfigure}

    \vspace{2mm}

    \begin{subfigure}[b]{0.49\linewidth}
        \centering
        \includegraphics[width=\linewidth]{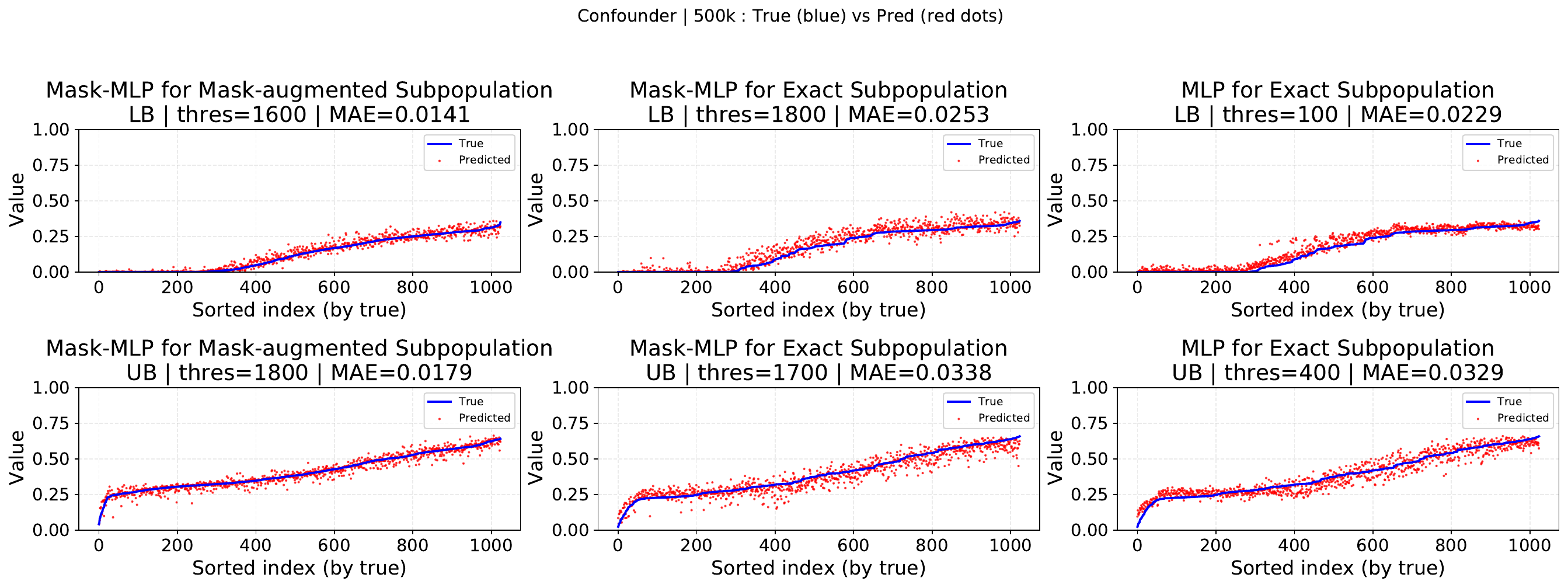}
        \caption{Confounder}
        \label{fig:app_tp_conf_500k}
    \end{subfigure}
    \hfill
    \begin{subfigure}[b]{0.49\linewidth}
        \centering
        \includegraphics[width=\linewidth]{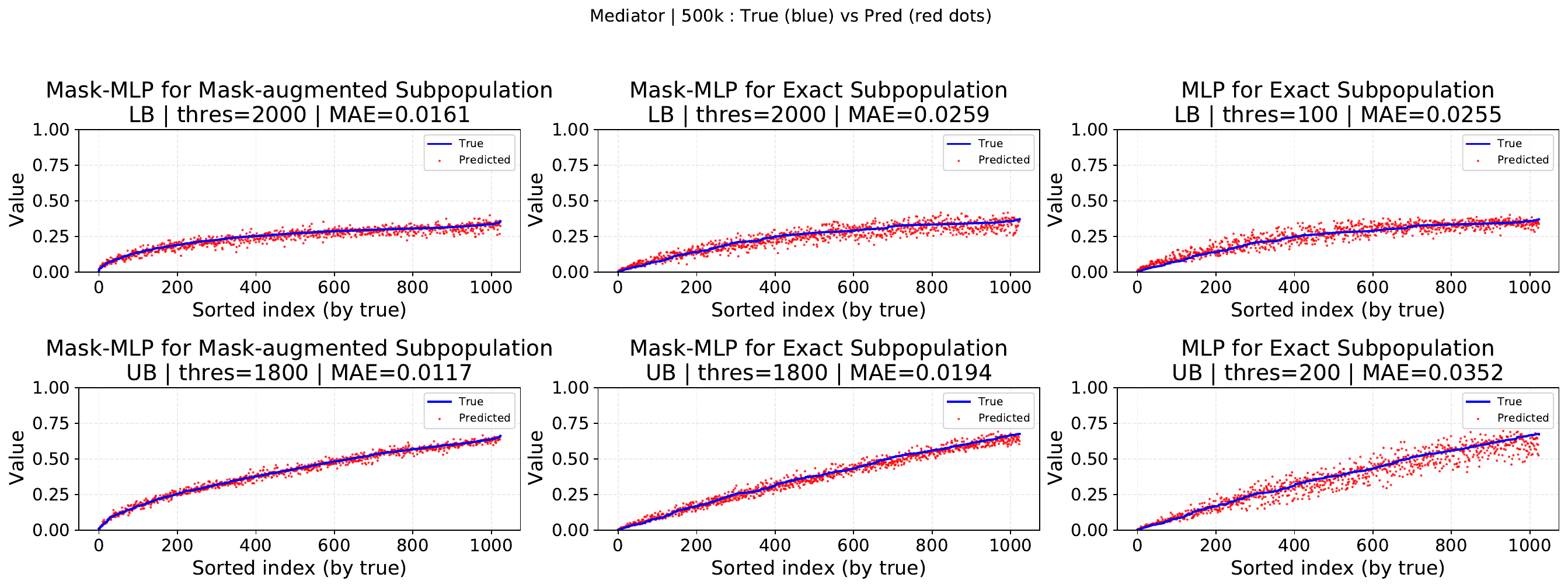}
        \caption{Mediator}
        \label{fig:app_tp_med_500k}
    \end{subfigure}
    \caption{True vs.\ predicted oracle bounds (500k budget).}
    \label{fig:app_truepred_500k}
\end{figure}

\end{document}